\documentclass[acmlarge]{acmart}
\AtBeginDocument{%
  }

\usepackage{subcaption}
\usepackage{makecell}
\usepackage{multirow}
\usepackage{longtable}
\usepackage{balance}
\usepackage{tabularray}
\usepackage{svg}
\usepackage{stfloats}

\setcopyright{none}
\acmConference{}{}{}
\acmBooktitle{}
\acmDOI{} 
\acmISBN{} 
\makeatletter
\renewcommand{\@authorsaddresses}{}
\renewcommand\footnotetextcopyrightpermission[1]{} 
\makeatother

\begin{document}

\title{Queryable 3D Scene Representation: A Multi-Modal Framework for Semantic Reasoning and Robotic Task Planning}

\author{Xun Li}
\orcid{0000-0002-1717-0669}
\affiliation{%
  \institution{CSIRO}
  \city{Sydney}
  \state{NSW}
  \country{Australia}
}
\email{xun.li@csiro.au}

\author{Rodrigo Santa Cruz}
\orcid{0000-0002-5273-7296}
\affiliation{%
  \institution{CSIRO}
  \city{Pullenvale}
  \state{QLD}
  \country{Australia}
}
\email{rfsantacruz@gmail.com}

\author{Mingze Xi}
\orcid{0000-0003-1291-4136}
\affiliation{%
  \institution{CSIRO}
  \city{Canberra}
  \state{ACT}
  \country{Australia}
}
\email{mingze.xi@csiro.au}

\author{Hu Zhang}
\orcid{0009-0009-9892-9515}
\affiliation{%
  \institution{CSIRO}
  \city{Sydney}
  \state{NSW}
  \country{Australia}
}
\email{hu1.zhang@csiro.au}

\author{Madhawa Perera}
\orcid{0009-0009-0928-0380}
\affiliation{%
  \institution{CSIRO}
  \city{Canberra}
  \state{ACT}
  \country{Australia}
}
\email{madhawa.perera@csiro.au}

\author{Ziwei Wang}
\orcid{0000-0002-0107-7347}
\affiliation{%
  \institution{CSIRO}
  \city{Pullenvale}
  \state{QLD}
  \country{Australia}
}
\email{ziwei.wang@csiro.au}

\author{Ahalya Ravendran}
\orcid{0000-0002-0107-7347}
\affiliation{%
  \institution{CSIRO}
  \city{Sydney}
  \state{NSW}
  \country{Australia}
}
\email{ahalya.ravendran@csiro.au}

\author{Brandon J. Matthews}
\orcid{0000-0002-8673-2434}
\affiliation{%
  \institution{CSIRO}
  \city{Canberra}
  \state{ACT}
  \country{Australia}
}
\email{brandon.matthews@csiro.au}

\author{Feng Xu}
\orcid{0000-0002-3932-1093}
\affiliation{%
  \institution{CSIRO}
  \city{Sydney}
  \state{NSW}
  \country{Australia}
}
\email{feng.xu@csiro.au}

\author{Matt Adcock}
\orcid{0000-0001-6191-9887}
\affiliation{%
  \institution{CSIRO}
  \city{Canberra}
  \state{ACT}
  \country{Australia}
}
\email{matt.adcock@csiro.au}

\author{Dadong Wang}
\orcid{0000-0003-0409-2259}
\affiliation{%
  \institution{CSIRO}
  \city{Sydney}
  \state{NSW}
  \country{Australia}
}
\email{dadong.wang@csiro.au}

\author{Jiajun Liu}
\orcid{0000-0001-8160-1796}
\affiliation{%
  \institution{CSIRO}
  \city{Pullenvale}
  \state{QLD}
  \country{Australia}
}
\email{jiajun.liu@csiro.au}

\renewcommand{\shortauthors}{Li et al.}

\begin{abstract}
To enable robots to comprehend high-level human instructions and perform complex tasks, a key challenge lies in achieving comprehensive scene understanding: interpreting and interacting with the 3D environment in a meaningful way. This requires a smart map that fuses accurate geometric structure with rich, human-understandable semantics.
To address this, we introduce the 3D Queryable Scene Representation (3D QSR), a novel framework built on multimedia data that unifies three complementary 3D representations: (1) 3D-consistent novel view rendering and segmentation from panoptic reconstruction, (2) precise geometry from 3D point clouds, and (3) structured, scalable organization via 3D scene graphs. Built on an object-centric design, the framework integrates with large vision-language models to enable semantic queryability by linking multimodal object embeddings, and supporting object-level retrieval of geometric, visual, and semantic information. The retrieved data are then loaded into a robotic task planner for downstream execution.

We evaluate our approach through simulated robotic task planning scenarios in Unity, guided by abstract language instructions and using the indoor public dataset Replica. Furthermore, we apply it in a digital duplicate of a real wet lab environment to test QSR-supported robotic task planning for emergency response. The results demonstrate the framework's ability to facilitate scene understanding and integrate spatial and semantic reasoning, effectively translating high-level human instructions into precise robotic task planning in complex 3D environments.
\end{abstract}

\keywords{3D Scene Representation, Multi-modal Scene Understanding, Natural Language Grounding, Robotic Task Planning, Multi-view captioning,Point cloud segmentation, Vision-Language Models, Unity}

\maketitle

\section{Introduction}
For robots to perform complex tasks in 3D environments under human instruction, they must relate high-level semantics in natural language commands to actual content in their surrounding environment. Even a simple instruction such as ``Robot, I’m thirsty.'' demands the ability to infer intent, locate relevant items (e.g., a water bottle), assess affordances, and plan a path for retrieval. Although trivial for humans, the tasks are exceptionally challenging for robots, as they must simultaneously reason about the spatial structure and semantic meaning of the environment based on human queries. While it is essential to enhance the robot's intelligence for navigating and manipulating complex environments, it is equally important to make the environment more understandable. We address this dual necessity by introducing the concept of a \textbf{queryable 3D scene representation (QSR)}, which embeds intelligence directly into the scene. This enables both robots and humans to interact with their surroundings in a more collaborative, context-aware, and semantically grounded manner.

Traditional 3D maps for robotic systems, such as voxel-based occupancy grids, point clouds, and mesh models~\cite{voxel_grids, point_cloud_classic, mesh_models, bandyopadhyay2024}, are predominantly geometric and often constructed using SLAM algorithms~\cite{slam1, slam2}. However, these representations lack the semantic information necessary for understanding and interacting with the scene. Semantic understanding, on the other hand, is typically derived from 2D object detection/segmentation models. The central challenge is aligning 2D semantics with 3D geometry to form a unified representation that enables complex reasoning and interaction. Moreover, human queries often span multiple levels of granularity and conceptual domains (e.g., ``a pillow with a tree pattern''), requiring far richer semantics than conventional models can provide.
Finally, human understanding of environments is inherently structural, involving hierarchical organisation and inter-object relationships. Capturing and reflecting the structural organisation in the map is essential for enhancing analytical capabilities and enabling more intuitive interaction with complex environments.

To address these challenges, we introduce 3D QSR, a scene-understanding multimodal framework built using multimedia data. It combines state-of-the-art 3D reconstruction techniques, such as NeRF~\cite{mildenhall2021nerf} and point clouds, with advanced semantic understanding through panoptic segmentation and vision-language embeddings. We also incorporate a 3D scene graph as an abstract layer, providing a structured, explicit, and lightweight representation enriched with object properties and inter-object relationships. Unlike single-modality systems, 3D QSR supports object-level queries involving location, appearance, function, and relational context, significantly enhancing scene understanding and interaction.
With a Large Language Model (LLM), our framework supports advanced language querying and reasoning grounded in QSR content. We demonstrate the capability of this representation through various downstream robotic task planning scenarios simulated in Unity~\cite{juliani2018unity} using the Replica dataset~\cite{straub2019replica}. In summary, the 3D QSR framework provides:

\begin{itemize}
\item Unified alignment of semantic, geometric, and structural information, enabling robots to reason over spatial and semantic context simultaneously.
\item Natural language-driven interaction, supporting intuitive query-answering for object retrieval.
\item Comprehensive support for robotic task planning, such as autonomous navigation, object retrieval, and adaptive decision-making in complex scenarios.
\end{itemize}

\section{Related Work}

\subsection{Aligning Semantics with Geometry}
As semantic information is primarily extracted from 2D visual representations, a key challenge lies in aligning 2D semantics with 3D geometry. A common approach is leveraging pretrained 2D object detection and semantic segmentation models to build semantic maps and project them into 3D space. For example, in~\cite{chaplot2020object}, the authors use differentiable projection operations to learn semantic mapping with supervision in the map space. SemanticFusion~\cite{mccormac2017semanticfusion} integrates geometry from SLAM with CNN-based semantic segmentation, while PanopticFusion~\cite{narita2019panopticfusion} combines predicted panoptic labels with volumetric maps, maintaining instance ID consistency.

With advances in 3D reconstruction methods such as NeRF~\cite{mildenhall2021nerf} and Gaussian Splatting~\cite{kerbl2023gaussian}, researchers have begun integrating semantics into these techniques to form semantic-aware neural scene representations. Notable works include Panoptic Neural Fields (PNF)~\cite{kundu2022panoptic}, Panoptic Lifting~\cite{siddiqui2023panoptic}, and PLGS~\cite{wang2024plgs}. These approaches create a unified, multi-view consistent 3D panoptic reconstruction that enables novel view synthesis with panoptic segmentation, highly beneficial for robotic tasks. However, these methods are constrained to preselected object classes and exhibit limited flexibility in handling human instructions due to reliance on fixed class names. Furthermore, the methods have yet to be tested in robotic navigation and manipulation tasks.

Recent breakthroughs in large vision-language models (LVLMs), such as CLIP~\cite{radford2021learning} and BLIP~\cite{li2022blip}, have enabled open-vocabulary semantic understanding by pretraining on large-scale image-language association pairs. By representing natural language semantics in maps through latent representations, these models facilitate natural interaction with human instructions and enable the generation of queryable scene representations for robotic tasks.

\subsection{Queryable Maps for Robotic Task Planning}
Queryable scene representations (maps) have shown strong potential in robotic applications. With the advent of LVLMs, one straightforward approach is to create maps directly from 2D images and their semantic interpretations. For instance, LM-Nav~\cite{shah2023lmnav} adopts this approach for long-horizon navigation in complex outdoor environments guided by natural language. Similarly, NLMap~\cite{chen2023nlmap} proposes an open-vocabulary, queryable semantic representation using ViLD~\cite{gu2021vild} and CLIP~\cite{radford2021learning}, integrating it into a language-based planner to address tasks beyond the reach of traditional planners. However, these methods often overlook the underlying geometry and structural representations of scene objects.

One improvement is fusing pretrained vision-language features extracted from LVLMs with traditional 3D representations like meshes and point clouds. VLMaps~\cite{huang2023visual} computes dense pixel-level embeddings using LVLMs from video feeds and back-projects them onto 3D surface maps. ConceptFusion~\cite{jatavallabhula2023conceptfusion} integrates features from foundation models into detailed 3D maps of unordered points, enabling zero-shot querying without retraining. These methods align visual-language features with 3D geometry, facilitating semantic localisation of landmarks in 3D space. Such maps can be converted to grid or occupancy maps commonly used in robotics. Their primary limitation lies in the quality of 2D-3D alignment and the granularity of the underlying 3D representation.

Recent research also explores integrating learnable 3D representations, such as NeRF~\cite{mildenhall2021nerf}, with LVLMs to construct such maps. CLIP-Fields~\cite{shafiullah2022clipfields} maps spatial locations to visual and semantic embeddings, serving as spatial-semantic memory for robots. LERF~\cite{kerr2023lerf} trains a neural field via knowledge distillation from multi-scale CLIP and DINO features~\cite{Caron_2021_ICCV}. These approaches combine the advantages of NeRF and semantic embeddings, enabling photo-realistic novel view rendering from query input—an essential feature for preview tasks. However, they are often constrained to specific scenes.

A novel abstraction, 3D Scene Graphs, has emerged as a structural representation for 3D environments. In~\cite{gu2024conceptgraphs}, the authors proposed an open-vocabulary graph-structured 3D scene representation and demonstrated its utility in robotic planning. Scene Graphs excel in providing structural, scalable representations with flexible levels of detail, integrating object relationships as graph edges. However, they lack detailed visual and geometric information.

The features offered by the three groups of LVLM-3D integrations are highly valuable and complementary. Our 3D QSR framework harnesses their respective strengths to create a unified, query\-able 3D representation while mitigating their respective limitations.

\begin{figure*}[t!]
    \centering
    \includegraphics[width=0.99\textwidth]{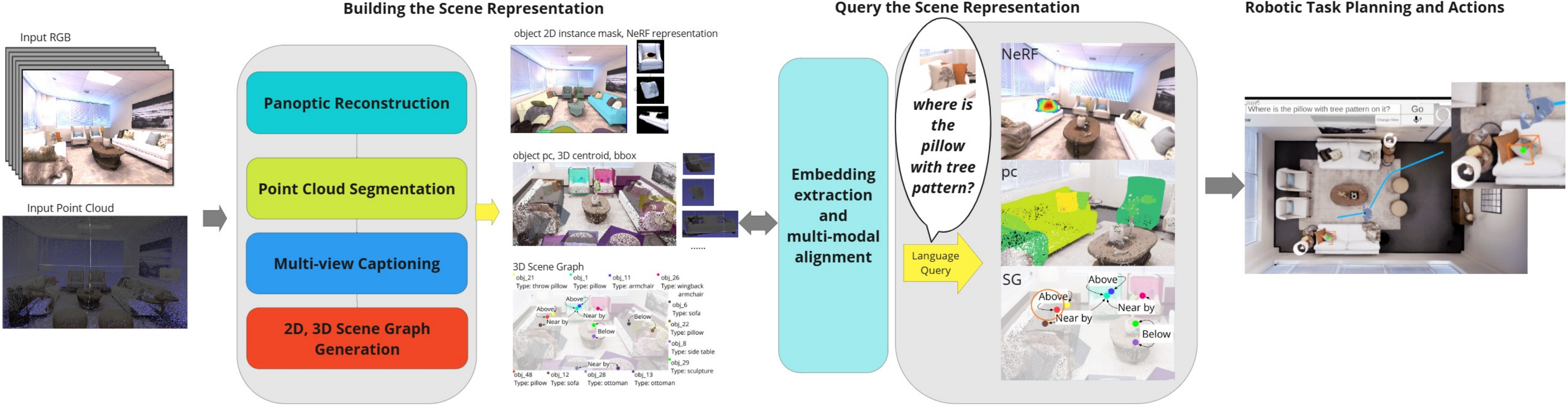}

    \caption{Three main parts of the workflow are: 3D QSR construction, 3D QSR query, and robotic task planning and action.}
    \label{fig:workflow}
    \Description{}

\end{figure*}

\begin{figure}[t]
	\centering\includegraphics[width=0.65\linewidth]{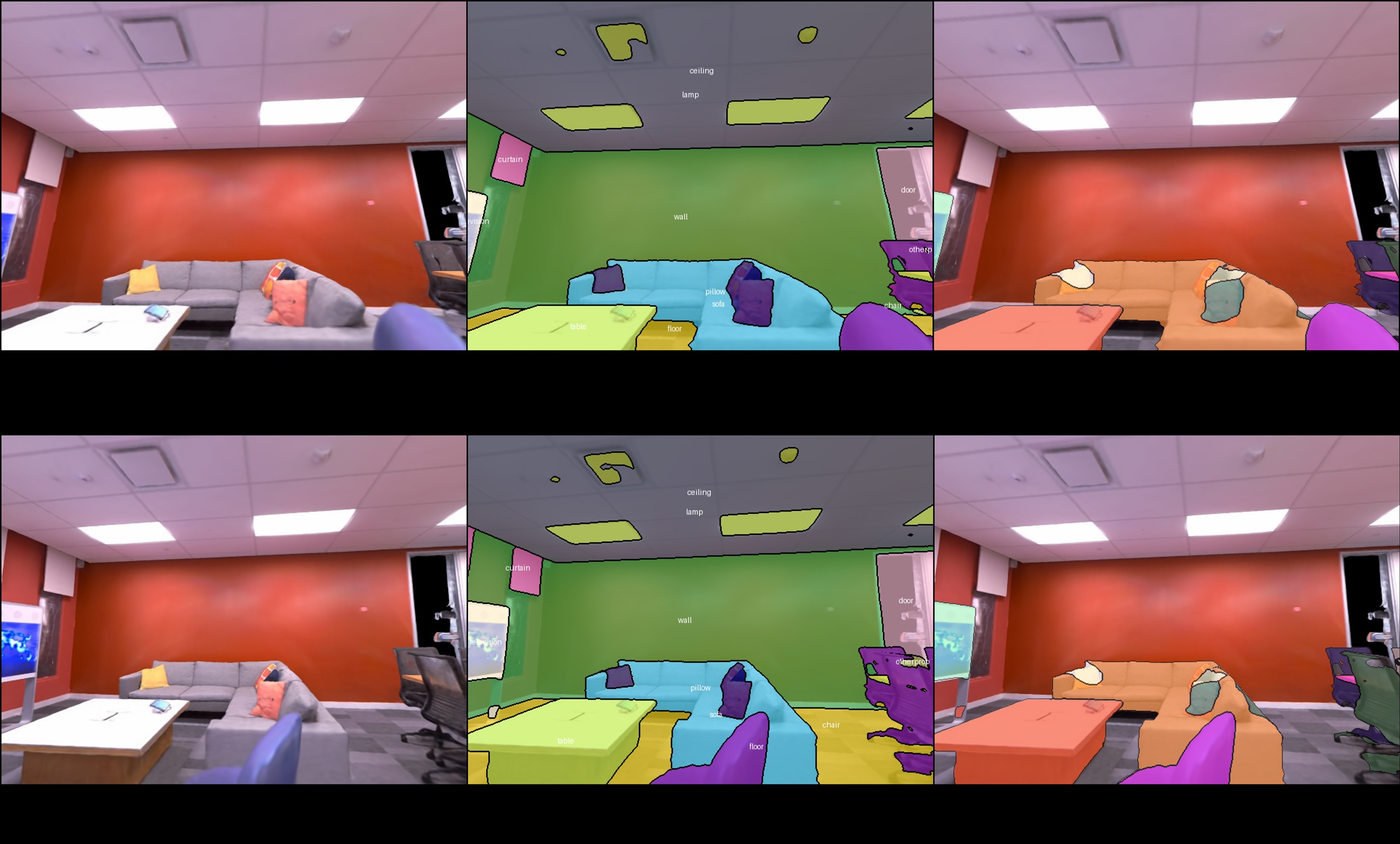}
	\caption{Consistent semantic and instance predictions across different frames.}    
	\label{fig: panoptic}
    \Description{}
\end{figure}

\section{Methodology and System Description}
\sloppy{Our framework constructs a unified, semantically rich and geometrically precise 3D representation by integrating three complementary elements: 1) panoptic radiance fields for consistent 3D reconstruction and segmentation, 2) segmented point clouds providing precise geometry, and 3) structured 3D scene graphs capturing object properties and relationships.} 
This enables object-level queries for robotic task planning, such as object localisation, property analysis, photorealistic view rendering, and spatial reasoning.

The framework accepts RGB videos and optionally point cloud data from sensors like LiDAR or generated from the video via Structure-from-Motion (SfM) techniques. First, Panoptic Lifting (PL)~\cite{siddiqui2023panoptic} generates a NeRF-based representation along with 3D-consistent semantic and instance segmentation. These 2D labels are then projected onto 3D point clouds to achieve geometric segmentation. These identified objects become nodes in a 3D scene graph, where an LVLM generates descriptive captions and infers relationships. 
By assigning shared identifiers across these three representations, the framework ensures seamless cross-referencing between visual, geometric, and semantic data.

For query capabilities, each representation uses tailored embeddings to enable the retrieval of comprehensive object information, including locations, physical attributes, semantic details, rendered images, and contextual relationships.
Finally, the query results are directly used by a robotic task planner to support downstream applications, such as autonomous navigation, object retrieval, and adaptive decision-making in complex environments. The overall workflow is shown in Figure ~\ref{fig:workflow} and detailed in the rest of this section.

\subsection{Building the Scene Representation}
\label{label:build-scene-rep}
\subsubsection{3D Panoptic Reconstruction} \label{subsubsec:panoptic}
This module employs Panoptic Lifting ~\cite{siddiqui2023panoptic} to simultaneously reconstruct scenes using an implicit NeRF model and generate consistent 2D semantic and instance segmentation masks across multiple views. Firstly, Mask2Former~\cite{cheng2021mask2former}, pre-trained on the ADE20K dataset~\cite{zhou2017scene}, is used to extract the semantic mask $M_s$ and instance mask $M_i$ for each frame individually. 
Although Mask2Former predictions can be inconsistent across frames, Panoptic Lifting can still achieve reasonable cross-view consistency through joint reconstruction and segmentation. It lays the foundation for object-level 2D–3D alignment.

Specifically, Panoptic Lifting includes two heads:
a semantic head predicting masks for all objects, and an instance head providing unique instance IDs for ``thing'' objects, which have regular shapes and clear boundaries.
As shown in Figure~\ref{fig: panoptic}, PL predicts consistent masks even with inconsistent and noisy input.
For each ``thing'' object, we compute the IoU between its instance mask and the semantic masks of all objects. We then choose the semantic mask with the highest IoU and assign its class name to the ``thing'' object.

\subsubsection{3D Point Cloud Segmentation} \label{subsubsec:pcdseg}

This module labels each point in the point cloud with an object identifier, leveraging consistent 2D segmentation masks obtained from PL.
For each 3D point, we project it onto multiple segmented images and construct a histogram  $\mathbf{h}_i \in \mathbb{Z}^C$, where each entry $\mathbf{h}_i[k]$ counts the number of times point $i$ aligns with segmentation label $k \in [0, \ldots, C-1]$ across different views, where $C$ is the total number of segmented instances. To ensure accurate labelling, only images where the point is visible are considered. Visibility is determined by checking if the depth of the point matches the depth of its projection in the depth map. The final label $l_i$ for point $i$ is then determined using a majority-voting approach:  
$l_i = \mathrm{argmin}~\mathbf{h}_i[k]$, where $k \in [0, \ldots, C-1]$.

\begin{figure}[t]
    \centering
    \includegraphics[width=0.99\linewidth]{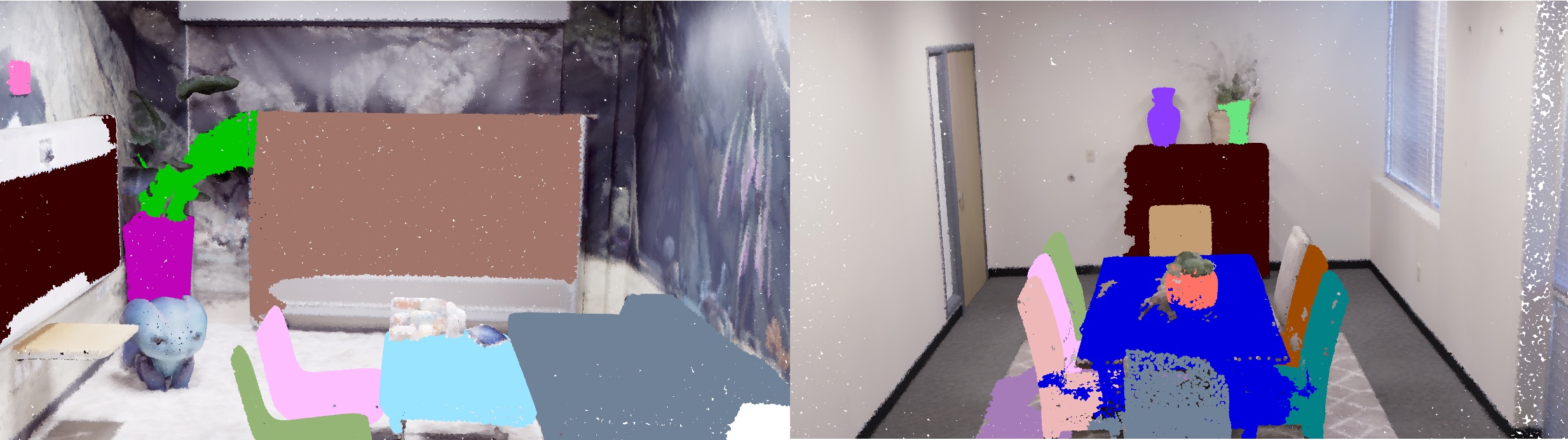}
    \caption{Point cloud object instance segmentation of two Replica scenes using the proposed algorithm.}
    \label{fig:segmentation_sample}
    \Description{}
\end{figure}

Points not visible in any view undergo a label propagation step and are assigned based on the majority label of neighbouring visible points within a predefined radius. 
DBSCAN is then applied to refine segmentation and remove noise, ensuring a tight and consistent representation of the object by retaining only its core points. Figure~\ref{fig:segmentation_sample} presents segmentation results for two scenes in the Replica dataset.

\subsubsection{Multi-view Caption Generation} \label{subsubsec:captioning}
Following accurate 3D reconstruction and segmentation, multi-view caption generation leverages LVLMs to enrich the semantic understanding of identified objects. By synthesising multiple viewpoints, this method addresses single-view limitations, such as occlusion or partial visibility, to produce comprehensive and contextually accurate object descriptions. This significantly reduces ambiguity and enhances descriptive fidelity for downstream analyses.

This methodology employs a two-stage approach. In the first stage, GPT-4o and Molmo-7B-D-0924~\cite{molmo2024} generate per-frame segmentation captions, capturing fine-grained segment details. The prompt focuses exclusively on the target object and explicitly avoids background elements and irrelevant details. While an ADE20K class label (from Panoptic Lifting) provides guidance, the model remains skeptical of potential misclassifications, especially for small or occluded objects. Structured prompts ensure consistency by covering key attributes like type, colour, material, shape, and function, with stylistic constraints and examples. If multiple objects are present, the model makes an informed selection.

In the second stage, GPT-4o synthesises these segment-level captions into a unified, coherent description. This step resolves discrepancies across different views, removes redundancy, and maintains objectivity by avoiding subjective or aesthetic commentary, ensuring the final caption is factual, definitive, and free of speculation.

\begin{figure}[t]
    \centering
    \begin{subfigure}[b]{0.35\linewidth}
        \centering
        \includegraphics[width=0.72\linewidth]{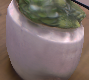}
        \caption{Bowl $\Rightarrow$ Vase}
        \label{fig:obj_vase}
    \end{subfigure}
    \hfill
    \begin{subfigure}[b]{0.62\linewidth}
        \centering
        \includegraphics[width=0.86\linewidth]{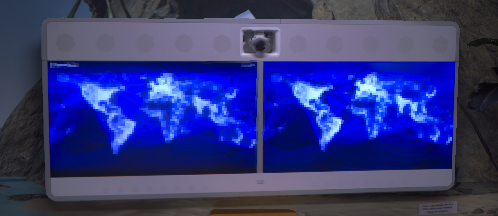}
        \caption{TV $\Rightarrow$ Video Conferencing System}
        \label{fig:obj_vcsystem}
    \end{subfigure}
    \caption{Object labels refined by multi-view captioning.}
    \label{fig:label_correction}
    \Description{}
\end{figure}

This approach notably improves object recognition accuracy by correcting erroneous labels from segmentation models, directly benefiting subsequent scene graph construction (see Section~\ref{sub-sec:sg-gen}). The following examples highlight cases where initial misclassifications were corrected through the multi-view approach. 
Examples in Figure~\ref{fig:label_correction} illustrate the effectiveness of multi-view captioning in rectifying errors, such as reclassifying a bowl as a vase (Figure~\ref{fig:obj_vase}) and a TV as a video conferencing system (Figure~\ref{fig:obj_vcsystem}).

We conducted a loose comparison to demonstrate that LVLMs can rectify mislabelled objects, reducing errors in scene graph construction. In this comparison, both “table” and “coffee table” are considered correct for a wooden coffee table. GPT-4o achieved 94.85\% accuracy, significantly surpassing Molmo-7B (77.32\%) and the Panoptic Lifting baseline (76.29\%). While Molmo-7B performs similarly to the baseline, it provides more granular labels, such as accurately identifying an “armchair” instead of the generic “chair” preset. In addition, attributes such as texture, colour, function, material, and pattern are extracted from the multi-view captions and saved using structured JSON format for downstream tasks.

Beyond correcting labels, the captions serve as foundations for detailed scene queries. 
Below is a snippet of a richly detailed multi-view caption of the video conferencing system shown in Figure~\ref{fig:obj_vcsystem}:

\textit{``This is a \textbf{dual-screen video conferencing system}... The device features \textbf{two large, rectangular screens placed side by side}, each \textbf{displaying vibrant digital maps} in shades of blue. It is housed in a sleek, predominantly \textbf{white frame}... At the top centre of the unit, there is \textbf{an integrated camera}... The device also incorporates \textbf{multiple circular speakers} aligned across the top edge... The system’s primary purpose is to facilitate virtual meetings and presentations...''}

\subsubsection{Scene Graph Generation}
\label{sub-sec:sg-gen}
\begin{figure}[h]
    \centering
    \includegraphics[width=0.7\linewidth]{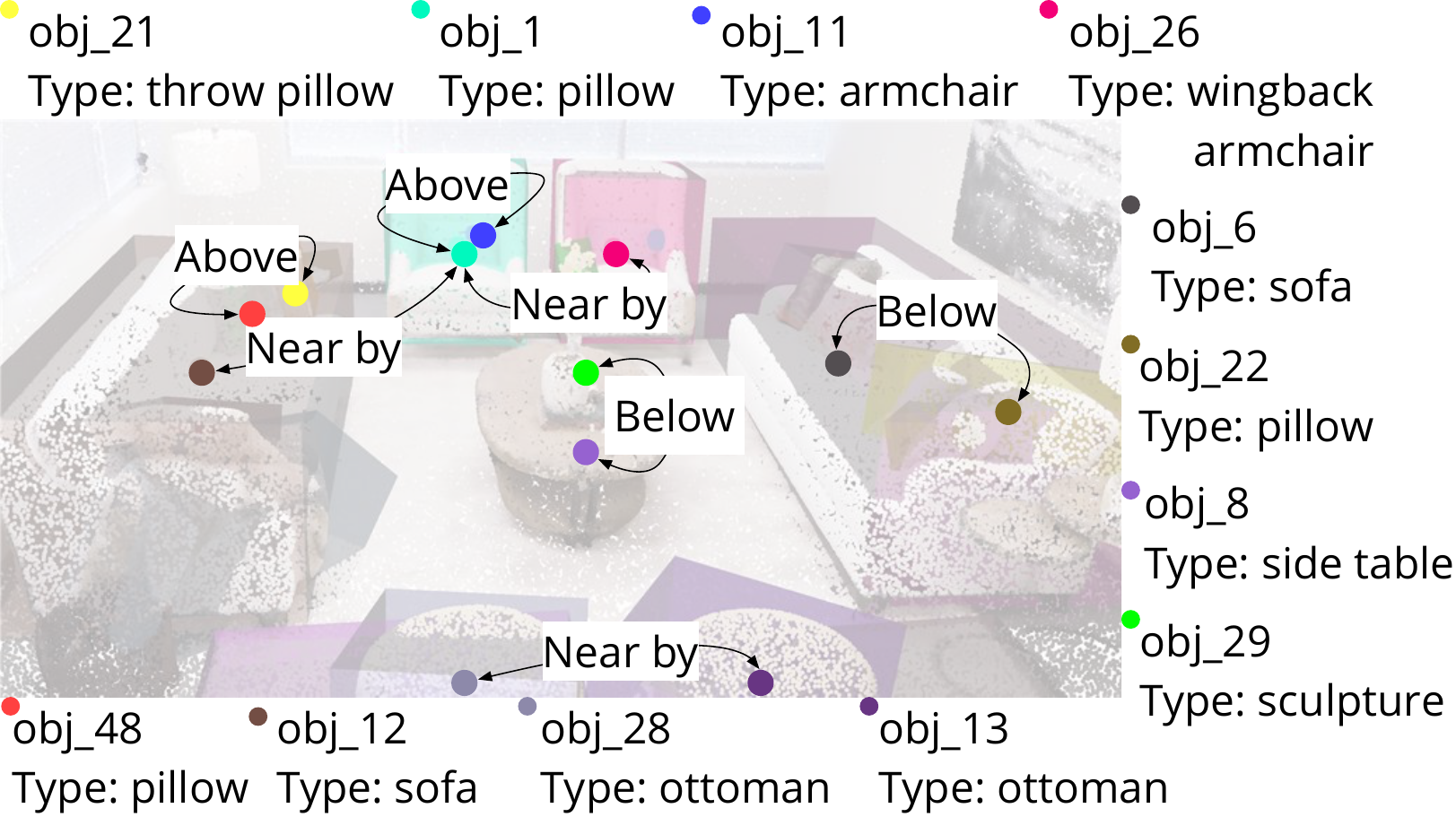} 
    \caption{A sample illustration of a generated 3D scene graph}
    \label{fig:scene graph}
    \Description{}
\end{figure}
Building upon the refined multi-view captions, this module constructs comprehensive 3D scene graphs by integrating geometric, semantic, and relational information. These graphs offer structured, contextual representations essential for advanced robotic perception and interaction.

Firstly, we generate 2D scene graphs at frame level, representing objects as nodes and their spatial or semantic relationships as edges.
These object nodes are instantiated using the consistent segmentation masks from Panoptic Lifting, enriched by detailed multi-view captions. 
Then, these objects are fed into the LLama3-8B model to infer their relationships.
To reduce computational costs, a filter is applied to only keep nearby objects for relationship inference.

Frame-level scene graphs are then aggregated to form a consistent 3D scene graph for the entire scene. During aggregation, 3D spatial properties—such as centroid positions and volume bounding boxes—along with object types, are refined based on the segmented points and captioning results. Object relational edges are preserved selectively based on spatial proximity criteria (e.g., centroids within one metre), ensuring the resulting scene graph maintains clarity and structural coherence. An illustrative example is shown in Figure~\ref{fig:scene graph}.

\subsection{Querying the Scene Representation}

\subsubsection{Querying 3D Point Cloud}
Robotic tasks often require object localisation in cluttered environments based on textual descriptions. Our approach enables this by leveraging LVLMs such as CLIP to embed segmented objects. For each object, we select ten views that observe most of its points, determined using depth consistency checks, as described earlier. A minimum-area bounding box is computed around the projected visible points, and the image crop is extracted from this bounding box. These crops are then enlarged by 20\% and 40\% for additional context and embedded by an LVLM.

\begin{figure}[t]
    \centering
    \includegraphics[trim=0 0 0 60, clip, width=0.65\linewidth]{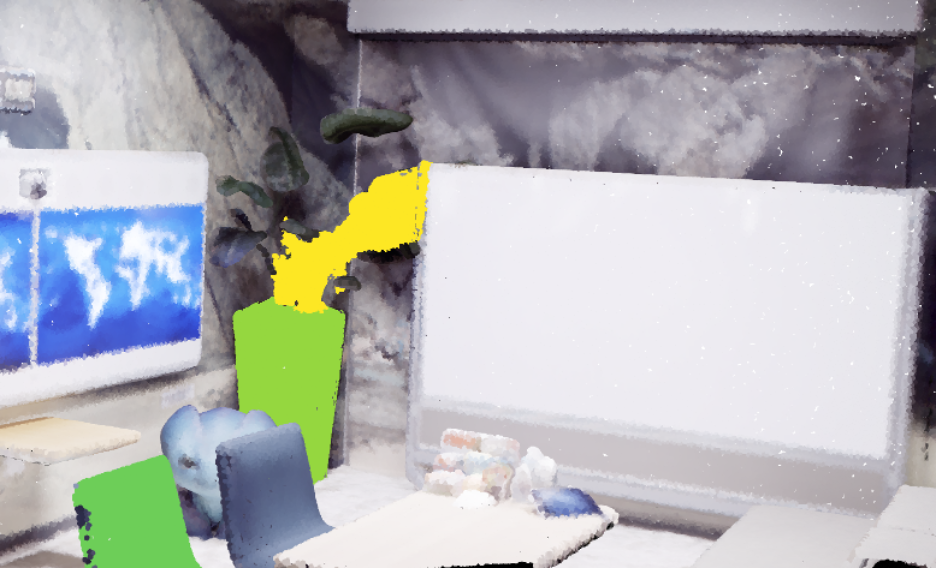}
    \caption{Plant decorations are retrieved and highlighted.}
    \label{fig:pc_query}
    \Description{}
\end{figure}

Textual queries undergo the same embedding process, and cosine similarity measures identify the objects most relevant to the queries. Figure~\ref{fig:pc_query} illustrates a successful retrieval result for the query "Any plant decoration in the room?".

\subsubsection{Querying NeRF}
Similarly, we leverage LERF~\cite{kerr2023lerf} for querying and visualising NeRF representations. LERF optimises a dense, multi-scale language-aligned 3D field using CLIP and DINO features~\cite{amir2021deep}.
This enables versatile open-vocabulary object localisation across cluttered environments.

\begin{figure}[t]
	\centering
	\includegraphics[trim=0 0 0 70, clip, width=0.9\linewidth]{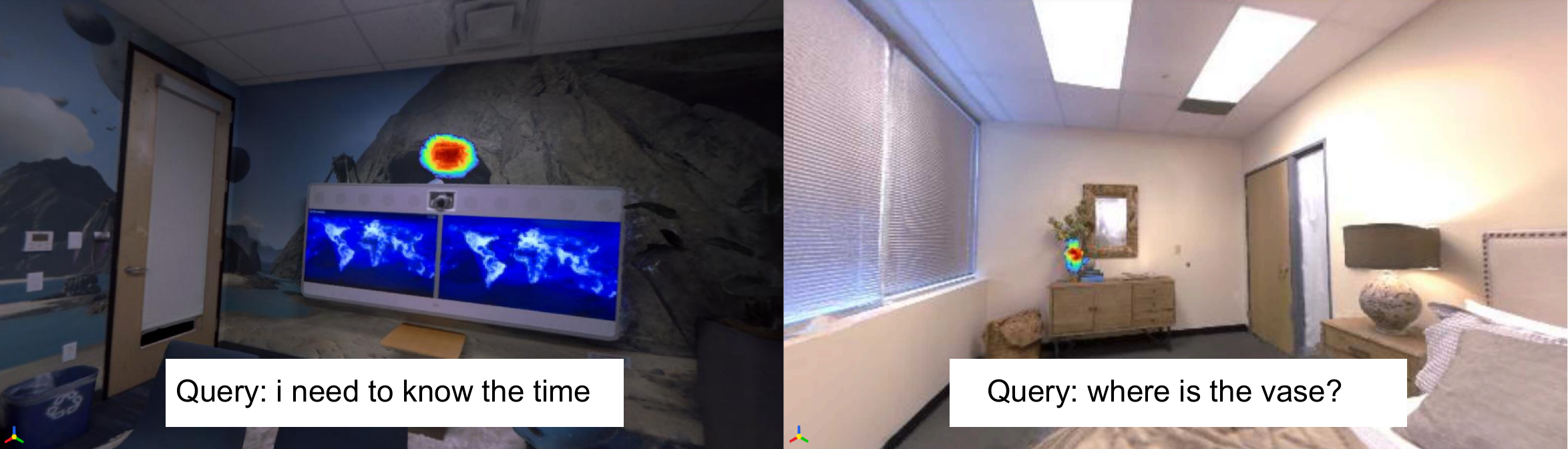}
\caption{Relevance maps of the NeRF with a descriptive query (left) and an affordance-based localisation query (right)}
	\label{fig_6}
    \Description{}
\end{figure}

We then use this to query NeRF scenes built from the Replica dataset, using both descriptive (e.g., ``Where is the vase?'') and affordance-based queries (e.g., “I need to know the time”). As shown in Figure~\ref{fig_6}, our system accurately identifies objects like vases and clocks, even without explicit queries, enabling semantic understanding and affordance-driven navigation in cluttered 3D environments.

\subsubsection{Querying Scene Graph}
\label{label:querying-sr}
Complementing point cloud and NeRF queries, we develop a scene graph (see Section~\ref{sub-sec:sg-gen}) querying module using retrieval-augmented generation (RAG) for effective natural language interaction with structured spatial data.
The generated SG, enriched by multi-view captions, includes object attributes (type, texture, colour, function) and spatial properties (centroids, bounding boxes), which are converted into semantic embeddings using the BAAI/bge-large-en-v1.5 model~\cite{bge_embedding}, which optimises retrieval quality by generating high-dimensional representations of textual data. The embeddings are indexed using FAISS (Facebook AI Similarity Search)~\cite{johnson2019billion} for efficient retrieval. When users submit natural language queries, embeddings are generated and similarity searches retrieve relevant scene nodes. The retrieved data are formatted for GPT-4o as contextual prompts to generate detailed, contextually accurate responses. This facilitates context-aware scene querying, enabling the extraction of spatial and semantic insights from modelled 3D environments.

\subsection{Linking Scene Understanding to Actions}
\label{label:connecting-robots}


\begin{figure}[t]
    \centering
    \includegraphics[width=0.8\linewidth]{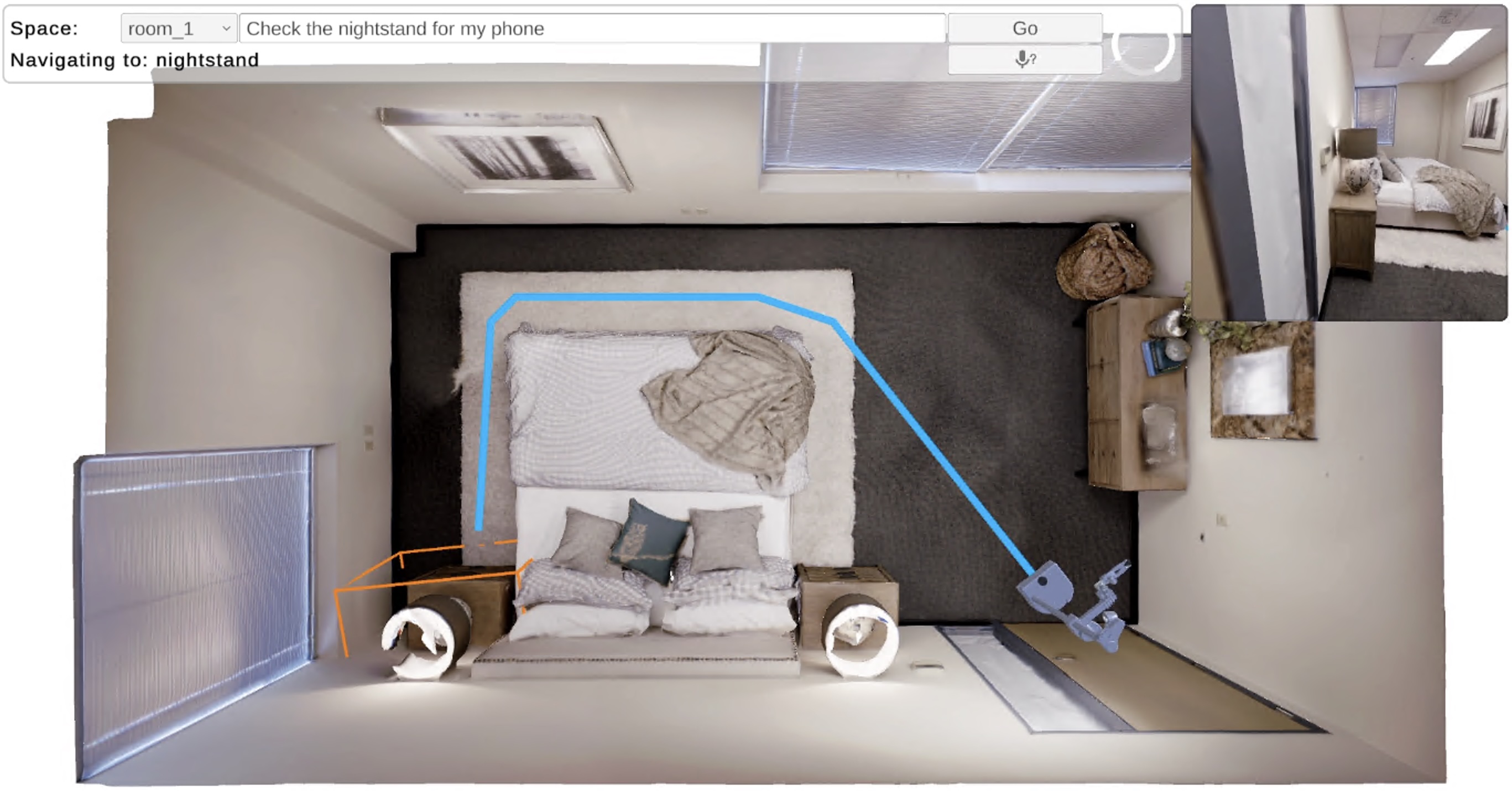}
    \caption{Visualisation of the computed robot navigation path to the object of interest, determined by the user query.}
    \label{fig:robot_nav}
    \Description{}
\end{figure}

For demonstration, we created a simulated environment in the Unity game engine using the Unity AI Navigation package, which can be treated as an ideal robotics navigation system. The URDF model of the Stretch 3 robot was included as an example in this simulation. For each Replica Dataset environment (room\_0, room\_1, room\_2, office\_1, office\_2, and office\_3) where the 3D QSR was developed, we imported a coloured mesh and generated a navigation mesh (NavMesh) in Unity. 
The 3D-QSR querying system was exposed via a RESTful API, responding to user queries with target object locations, centroids, and bounding boxes.
The Unity simulation renders the bounding box around the target object, computes and visualises a navigation path, as exemplified in Figure~\ref{fig:robot_nav}.
Then, the simulated Stretch 3 robot navigates to the target object. A camera attached to the head of the robot observes the target object during navigation. 
We discuss this demonstrator system briefly in Section~\ref{label:query-robot-nav} with some example queries. Overall this demonstration showcases how connecting 3D QSR with robot actions allows for natural and intuitive HRI, enabling spatially-aware and context-driven navigation and object retrieval tasks.

\section{System Demonstration and Evaluation}
We evaluate object retrieval performance across three modalities—point cloud, NeRF, and SG—across six Replica scenes. We discuss and further illustrate 3D QSR’s role in effective robotic navigation in Unity. Additionally, we present a preliminary exploration of integrating LLMs with real-time scene observations to refine task planning through iterative validation and reasoning.
~\footnote{Refer to Supplementary B for querying result demonstrations for PC, NeRF, and SG (visualised in the Unity simulator). Refer to Supplementary C for queries generated by an LLM for performance evaluation in Table 2. See the videos in the supplementary materials showcasing querying capabilities for robotic task planning.}

\subsection{Closed-set vs Open-vocabulary}
\begin{table}[t]
  \centering
  \caption{The comparison between Panoptic Lifting and the open-vocabulary reconstruction model. Both methods show competitive results in reconstructed images (PSNR); however, Panoptic Lifting obtains much better object masks (mIoU).}
  \tabcolsep=.4cm
  \renewcommand{\arraystretch}{0.2}
   \begin{tabular}{c|c|cc}
    \toprule
    Scene & Setting & PSNR~$\uparrow$   & mIoU~$\uparrow$ \\
    \midrule
    \multirow{2}{*}{Office\_0} 
     & open-vocabulary & 37.23 & 0.3409  \\
     & closed-set &  38.02   & 0.4345 \\
        \midrule
    \multirow{2}{*}{Office\_2}
     & open-vocabulary & 35.33  & 0.5590  \\
     & closed-set & 36.05 &  0.7691 \\
      \midrule
     \multirow{2}{*}{Office\_3}
     & open-vocabulary & 35.16   & 0.4696   \\
     & closed-set &34.80  & 0.6811  \\
      \midrule
     \multirow{2}{*}{Office\_4}
     & open-vocabulary & 34.10  & 0.5141  \\
     & closed-set & 33.97  & 0.6580  \\
      \midrule
     \multirow{2}{*}{Room\_0}
     & open-vocabulary & 32.67  & 0.4462  \\
     & closed-set & 32.53  &  0.6671 \\
      \midrule
     \multirow{2}{*}{Room\_1}
     & open-vocabulary & 33.48  &  0.5797 \\
     & closed-set & 33.31   & 0.5836   \\
      \midrule
     \multirow{2}{*}{Room\_2}
     & open-vocabulary & 35.06  & 0.5823  \\
     & closed-set &  35.56 &  0.8204 \\
    \bottomrule
    \end{tabular}
  \label{tab:pn and ov}%
\end{table}%

To assess the suitability of our panoptic reconstruction module, we compared closed-set Panoptic Lifting against our open-vocabulary variation, which is expected to exhibit better generalisation to unseen object categories. 
Specifically, we adapted OpenSeg~\cite{ghiasi2022scaling}, a 2D open-vocabulary segmentation model, to our reconstruction pipeline and evaluated it across seven Replica dataset scenes~\cite{straub2019replica} in Table~\ref{tab:pn and ov}. 
Despite the improved flexibility of open-vocabulary methods, the generated masks exhibit significantly lower quality, resulting in inferior reconstruction performance (e.g., lower mIoU). Consequently, we continue to use the original closed-set Panoptic Lifting in this work to fully exploit the capabilities of downstream modules, while leaving open vocabulary as future work.

\subsection{Evaluation of Querying Performance}
We evaluated the queryable capability across all three representations using descriptive, affordance, and negation queries. For descriptive queries, such as “cushions,” the system retrieves all relevant instances with centroids and bounding boxes. It also handles fine-grained semantics—e.g., “a cushion with a tree pattern”—and is robust to variations like replacing “cushion” with “pillow”
(Figure~\ref{fig:descriptive_spec}; full results in Supplementary B). 

\begin{figure*}[t!]
    \centering
    \includegraphics[width=0.99\textwidth]{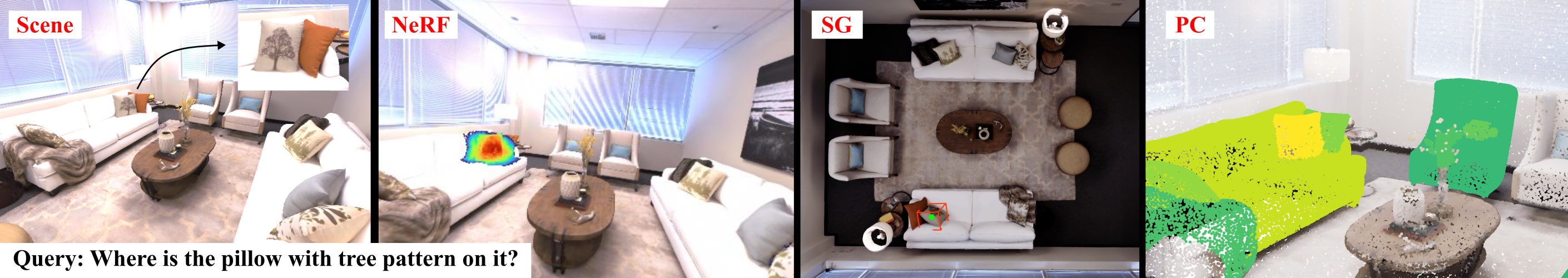}
    \caption{Retrieval results show the system can identify a highly specific query, ``a pillow with a tree pattern on it''.}
    \label{fig:descriptive_spec}
    \Description{}

\end{figure*}

\begin{figure*}[t!]
    \centering
    \includegraphics[width=0.99\textwidth]{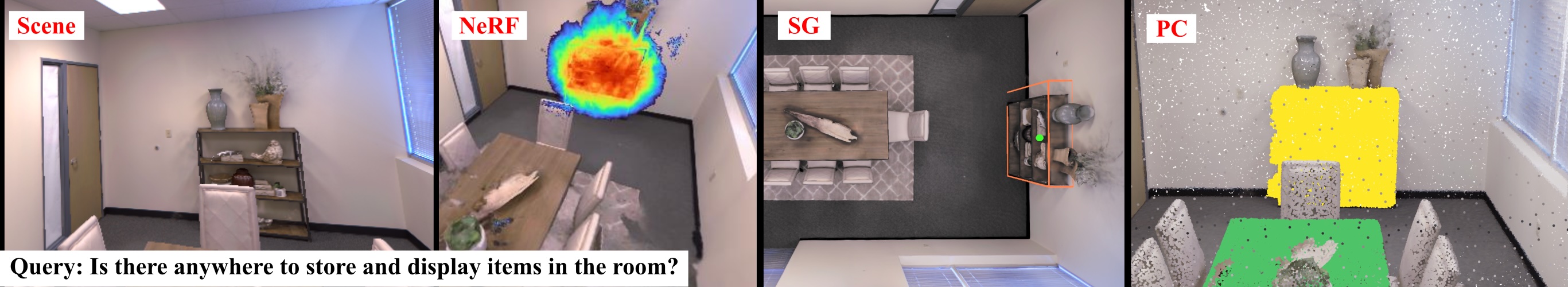}
    \caption{The system identifies objects based on latent information (e.g., ``Where to store and display items in the room?'').}
    \label{fig:descriptive_abstract}
    \Description{}
    
\end{figure*}

Affordance queries infer object functionality beyond appearance; for example, “Is there anywhere to store and display items?” correctly highlights shelving units (Figure~\ref{fig:descriptive_abstract}). 
Negation queries exclude specific categories, such as “Anything to sit on other than a chair?”, which retrieves a sofa but not a chair.

To systematically evaluate, we used GPT-4o to generate a set of 20 queries per scene: 10 descriptive, 5 affordance, and 5 negation  (full query list in Supplementary C). For descriptive queries, where the target object(s) are clearly identifiable by a human annotator, the performance was evaluated via precision and recall to show how accurately and completely the retrievals are.

In contrast, affordance and negation queries are inherently more open-ended and may correspond to multiple valid objects in a scene. For example, the query ``Something rigid, unlike a plush pillow.'' could apply to a large variety of objects in a room, making it infeasible for a human annotator to exhaust all correct answers. Thus, we adopt a soft binary evaluation approach: if any of the retrieved objects satisfies the query condition, the result is marked as a success. We then compute the success rate as the proportion of successful responses across all affordance or negation queries. This flexible evaluation allows us to assess system performance in open-ended and subjective retrieval tasks without requiring exhaustive ground truth annotations. The result is shown in in Table~\ref{quantitative}. 
\begin{table}[t]
\centering
\caption{Average performance metrics for LeRF, Point Cloud and Scene Graph, showing descriptive-query precision and recall alongside affordance and negation success rates.}

\refstepcounter{table}
\begin{tblr}{
  row{2} = {c},
  cell{1}{1} = {r=2}{c},
  cell{1}{2} = {c=2}{c},
  cell{1}{5} = {c},
  cell{3}{2} = {c},
  cell{3}{3} = {c},
  cell{3}{4} = {c},
  cell{3}{5} = {c},
  cell{4}{2} = {c},
  cell{4}{3} = {c},
  cell{4}{4} = {c},
  cell{4}{5} = {c},
  cell{5}{2} = {c},
  cell{5}{3} = {c},
  cell{5}{4} = {c},
  cell{5}{5} = {c},
  hline{1,3,6} = {-}{},
  hline{2} = {2-5}{},
}
Methods & Descriptive &  & Affordance & Negation\\
 & Precision & Recall & Success Rate & Success Rate\\
Radiance Field & 0.71 & 0.88 & 0.83 & 0.43\\
Point Cloud &  0.88 &  0.91 & 0.80  & 0.71\\
Scene Graph & 0.74  &  0.72 &  0.93 &  0.90\\ 
\end{tblr}

\label{quantitative}
\end{table}

\begin{figure*}[t!]
    \centering
    \includegraphics[width=0.99\textwidth]{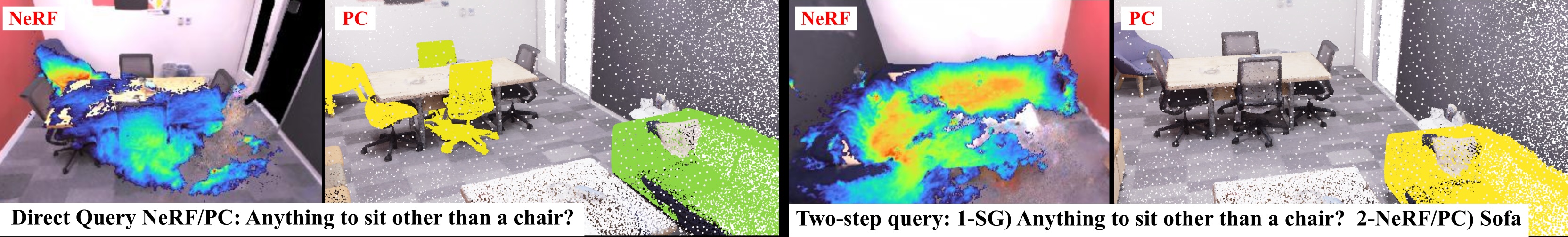}
    \caption{Demonstration of a two-step query approach, where the scene graph first processes complex queries (e.g., ``Anything to sit on other than a chair?'') and then precise information extracted from the scene graph (e.g., ``sofa'') are used for retrieval.}
    \label{fig:negation}
    \Description{}
    
\end{figure*}

While embeddings based methods (i.e., point cloud and NeRF query modules) excelled in descriptive queries, we observed that they struggle with more complex or nuanced queries, particularly the negation queries, where specific attributes or categories must be excluded. For example, a query like ``Anything to sit on other than a chair?'' tends to fail using such methods.
Addressing this limitation, we proposes a two-step query approach: initial query using SG, then passes retrieved entities to point cloud module and LeRF module. This leads to  improvement of the performance of 7\% for Affordance queries, and 50\% for Negation queries, with one example shown in Figure~\ref{fig:negation}. 

Finally, we highlight the different capabilities of the various queryable modules in our framework. Querying point clouds (PC) provides precise localisation and enables the computation of object geometric properties useful for downstream tasks (e.g., bounding boxes and centroids). The scene-graph-based query module provides additional property and relationship information, and also facilitates language-based high-level reasoning. 
NeRF-based querying provides novel view rendering for visualisation. 
Combining these three complementary approaches forms our framework, and we further discuss its application in the context of robot navigation tasks using Unity as a proof-of-concept (PoC) in Section~\ref{label:query-robot-nav}.

\subsection{Query-Driven Robot Navigation}
\label{label:query-robot-nav}

\begin{figure*}[t!]
    \centering
    \includegraphics[width=0.99\textwidth]{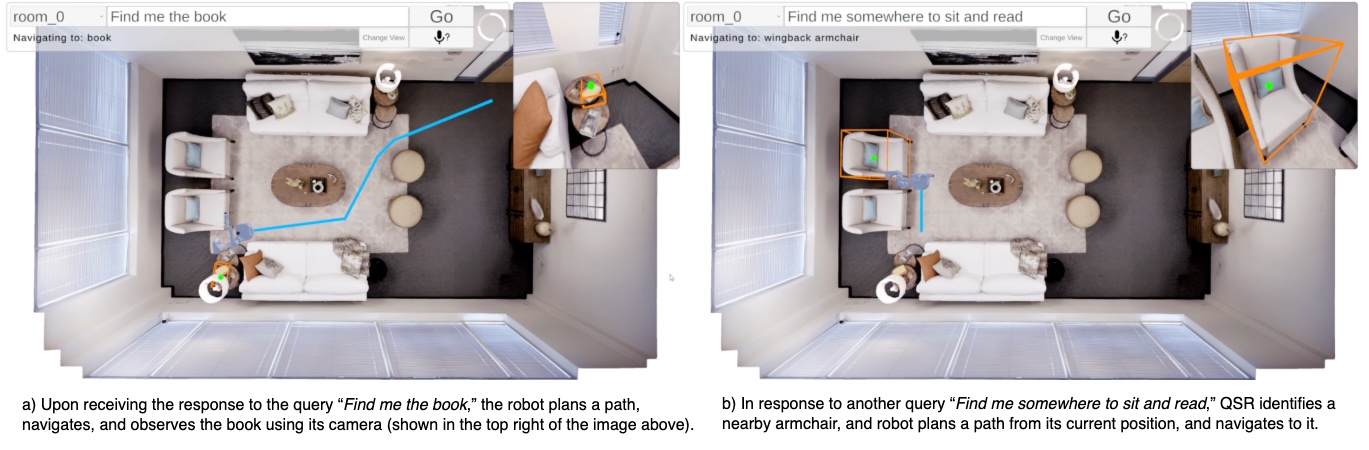}
    \caption{An example interaction with the QSR using the Unity simulation and querying interface. Using the results from querying the QSR, the simulated Stretch3 robot navigates room\_0 of the Replica dataset, navigating to the book and somewhere to sit and read. The video of this interaction is included in the supplemental material.}
    \label{fig:nav_video}
    \Description{}

    \includegraphics[width=0.99\textwidth]{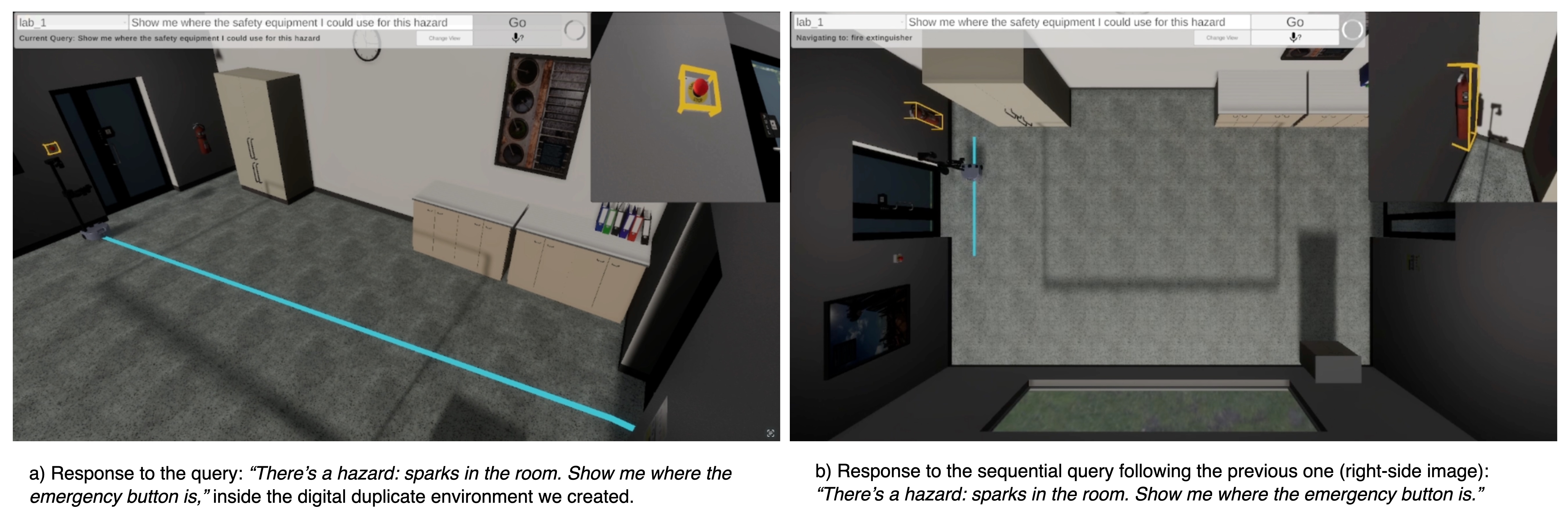}
    \caption{Query QSR of a digital duplicate of an actual wet lab for robot navigation}
    \label{fig:chemical_lab}
    \Description{}

\end{figure*}

We further validated our 3D QSR with robotic navigation tasks in our proof of concept (PoC) Unity simulator (see Section~\ref{label:connecting-robots}).
In this system, the user can submit a natural language query related to an object in the scene, as a written or spoken query (via an automatic speech recognition model). 
The system then uses the query capability of QSR to locate an object of interest, respond with its centroid and bounding box for path planning.

An example is illustrated in Figure~\ref{fig:nav_video}, where the robot responds to consecutive human queries: first locating a book, then finding a suitable place in the room to sit and read. As shown, the Stretch 3 robot navigation path (blue), target object centroid (green), and bounding box (orange) are visualised in the scene.

In addition to the Replica dataset environments, we used our own synthetic environment, a digital duplicate of an actual wet lab, as shown in Figure~\ref{fig:chemical_lab}, to further test our query-driven navigation performance. In this example, the robot is guided by a sequence of natural language queries: first, ``There’s a hazard: sparks in the room. Show me where the emergency button is,'' upon which the system locates the emergency button. This is followed by ``Show me where the safety equipment I could use for this hazard is,'' and the robot, by querying QSR, successfully identifies and navigates to a nearby fire extinguisher.

Furthermore, we can employ rendered NeRF as a simulator for task planning by generating photorealistic novel views of scenes corresponding to new poses of a robot or object, and the perspective of a simulated robot. Using a trained NeRF model, we generate accurate visual previews for task planners to infer how the environment appears from unexplored configurations. We can also take the pose of a simulated robot's camera, and generate the robot's novel perspective. This approach enables us to test and optimise task execution strategies in a simulated, photorealistic environment prior to physical deployment. The next section outlines how this approach supports more complex task planning.

\subsection{Preliminary Exploration: LLM-based scene consolidation and task planning with deep reasoning}
\label{deep-reasoning-poc}
A major advantage of using a comprehensive, pre-built 3D scene representation for task planning is that it offers a holistic view beyond the robot’s immediate observations. This allows the planner to reason about objects indirectly affected by an action, not just those explicitly mentioned in the task. But in real-world settings, environments change, and discrepancies between the pre-built representation and real-time observations are inevitable. This section addresses how to cross-validate and consolidate the scene representation with current observations to improve task reasoning.

\begin{figure*}[!htp]
	\centering
	\includegraphics[width=0.99\textwidth]{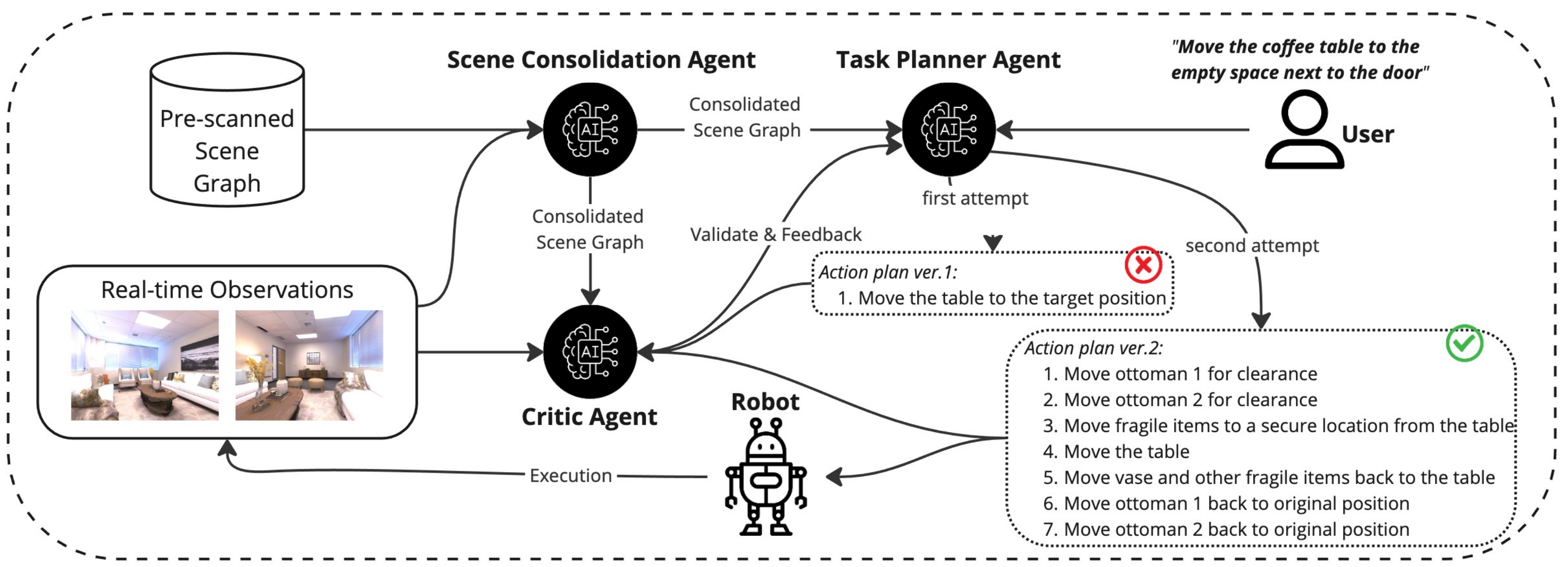}
	\caption{The process of LLM-based scene consolidation and task planning with deep reasoning.}
	\label{fig_7}
    \Description{}
\end{figure*}

To achieve this, we propose a real-time, interactive process that combines object- and relation-level information from both sources. As shown in Figure~\ref{fig_7}, a pre-scanned scene graph and real-time input are passed to a Scene Consolidation Agent, which updates the scene graph by detecting changes. The updated graph is then used by a Task Planner Agent to generate an action plan, which is evaluated by a Critic Agent for risks or inconsistencies. If accepted, the plan is executed; if rejected, it is revised with feedback.

For example, moving a coffee table should also consider fragile objects on top (like a vase), which require removal before movement. Similarly, the planner should clear the path (e.g., move ottomans) and restore them afterward. Such nuanced planning is only feasible when the environment is semantically and structurally represented in a form that both the Task Planner and Critic Agents can interpret.

\section{Limitations}
Despite its strengths, our 3D QSR framework has several limitations. Most notably, the current panoptic reconstruction module depends on a fixed-category closed-set segmentation model, which restricts the system’s ability to handle fully open-vocabulary queries. This constraint limits the number of object types that can be recognised and queried within a scene. Future work will explore open-vocabulary approaches that enhance generalisation while preserving high-quality reconstruction. Secondly, while effective in simulation, our framework has not been rigorously tested in real-world environments, which are more complex and unstructured. Deploying on real-world datasets will be essential for validating robustness. Thirdly, the current representation assumes a static environment, making it challenging to adapt to real-time object changes. Incorporating incremental scene updates with scene
consolidation will enhance real-world usability.

\section{Conclusion}
This work introduced 3D Queryable Scene Representation (3D QSR), a novel multimodal framework that unifies multiple 3D representations in an object-centric manner. By embedding pre-trained vision-language features within each representation, the framework enables query capabilities that allow simultaneous interpretation of visual, semantic, geometric, and structural aspects of the environment. 

Through simulated robotic task planning in Unity, we demonstrated how 3D QSR enables scene understanding, spatial reasoning, and high-level task execution from abstract human instructions. Our framework serves as a bridge between human intent and real-time robotic actions, enabling context-aware navigation, object retrieval, and informed decision-making. This system has demonstrated the initial potential of 3D QSR in enabling robots to interact with complex environments through natural language queries. Future work will focus on enhancing open-world recognition, adapting to dynamic environments, and validating in real-world settings. 

\balance

\bibliographystyle{ACM-Reference-Format}
\bibliography{2.reference}

\clearpage
\appendix




\section*{Supplemental Material: Queryable 3D Scene Representation: A Multi-Modal Framework for Semantic Reasoning and Robotic Task Planning}

\section{Sample Captioning Result}
\label{label:appendix-a}
\subsection{VC example}

This is a \textbf{dual-screen video conferencing system} designed for professional settings. The device features two large, rectangular screens placed side by side, each \textbf{displaying vibrant digital maps} in shades of blue. It is housed in a sleek, predominantly \textbf{white frame}. The frame is crafted from a combination of high-quality plastic and metal materials, ensuring durability and a polished appearance. At the top centre of the unit, there is \textbf{an integrated camera}, essential for capturing high-quality video during conferences. The camera may be housed in a slightly raised or metallic enclosure. The device also incorporates \textbf{multiple circular speaker} elements aligned across the top edge, providing audio output that complements its visual capabilities. The system primary purpose is to facilitate virtual meetings and presentations, providing seamless integration of video and audio components for effective communication. The overall design is minimalist and modern, ensuring it blends well into contemporary office environments.
\begin{figure}[!h]
    \centering
    \includegraphics[width=0.7\linewidth]{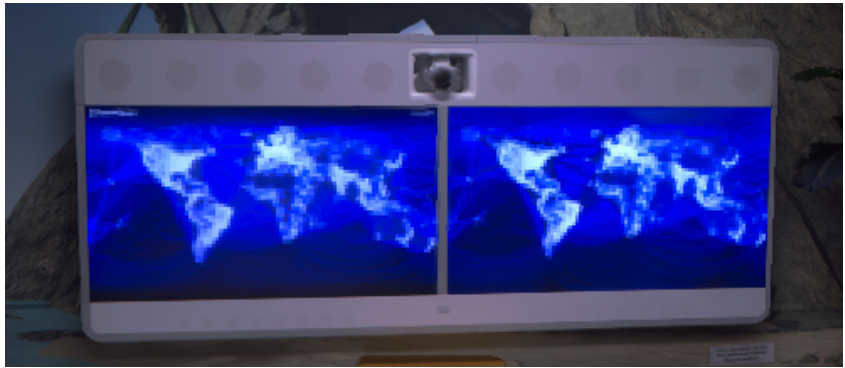}
    \caption{A video conference system from Replica Dataset.}
    \label{app-1}
    \Description{}
\end{figure}

\subsection{Arm Chair}
This modern \textbf{lounge chair} features a sleek, curved design with a single seat cushion and a backrest, both upholstered in light beige fabric. The chair includes \textbf{a light blue pillow with gold trim} placed against the backrest for added comfort. The design incorporates gracefully \textbf{curved armrests} that are seamlessly integrated into the structure, forming a continuous line from the base to the top of the backrest, creating a streamlined silhouette. The chair's geometry is elegant and minimalist, presenting a slightly reclined posture for comfortable seating. Its robust structure is \textbf{supported by a light-coloured wood frame\footnote{The wood frame is not visible from this view.}} with a visible grain pattern, potentially birch, which contributes to its sophisticated appearance. Suitable for indoor use, this chair is designed for comfort and versatility, making it an ideal addition to living rooms, waiting areas, or as a statement piece in contemporary interiors. The \textbf{neutral colour} palette and minimalist design ensure compatibility with a variety of decor styles, while its armchair-like shape offers a wide, inviting base that narrows towards the top, enhancing its elegant aesthetic.
\begin{figure}[!h]
    \centering
    \includegraphics[width=0.3\linewidth]{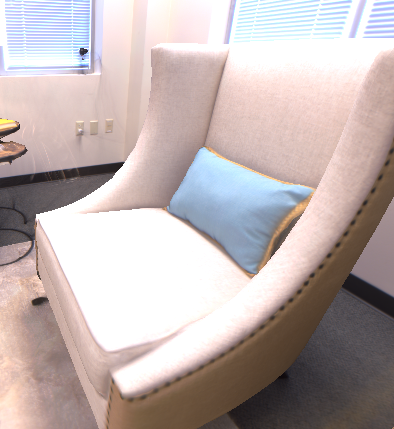}
    \caption{An arm chair from Replica dataset.}
    \label{armchair}
    \Description{}
\end{figure}

\subsection{Pillow}
This is a \textbf{decorative throw pillow} with a \textbf{soft cream base colour}, adorned with an \textbf{abstract pattern in earthy shades of green and brown}, featuring organic \textbf{foliage-inspired shapes}. The material is a high-quality, smooth, and soft fabric, likely a cotton blend, providing comfort and durability. The pillow has a classic \textbf{square shape}, enhanced with generous, plush cushioning, offering a cushy and inviting texture. Its primary Affordance is to serve as an accent piece, adding aesthetic appeal and providing additional comfort to seating areas such as sofas, beds, or lounge chairs.
\begin{figure}[!h]
    \centering
    \includegraphics[width=0.3\linewidth]{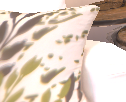}
    \caption{A pillow from Replica dataset.}
    \label{pillow}
    \Description{}
\end{figure}

\section{Query Result from three Modules}
\label{label:appendix-b}

\onecolumn
\begin{center}
\footnotesize
\setlength{\tabcolsep}{1pt}
\begin{longtable}{m{0.9cm}|m{1.3cm}|m{1.4cm}|m{1.3cm}|m{3cm} m{3cm} m{4cm}}

    \caption{Query Result Demonstration} \\
\toprule     
\makecell{\textbf{Room}} & \makecell{\textbf{Query} \\ \textbf{Type}}& \makecell{\textbf{Replica} \\ \textbf{Dataset} \\ \textbf{Semantic} \\ \textbf{Queries} \\ \textbf{for Room }\\\textbf{Scenes}} & \makecell{\textbf{Human} \\ \textbf{Annotated}\\ \textbf{Ground} \\ \textbf{Truth}}  & \makecell{\textbf{Query from} \\\textbf{Point Cloud Module}}& \makecell{\textbf{Query from} \\ \textbf{NeRF Module}}& \makecell{\textbf{Query from} \\ \textbf{Scene Graph Module}} \\ 
         \midrule
\endfirsthead
\toprule
\makecell{\textbf{Room}} & \makecell{\textbf{Query} \\ \textbf{Type}}& \makecell{\textbf{Replica} \\ \textbf{Dataset} \\ \textbf{Semantic} \\ \textbf{Queries} \\ \textbf{for Room }\\\textbf{Scenes}} & \makecell{\textbf{Human} \\ \textbf{Annotated}\\ \textbf{Ground} \\ \textbf{Truth}}  & \makecell{\textbf{Query from} \\\textbf{Point Cloud Module}}& \makecell{\textbf{Query from} \\ \textbf{NeRF Module}}& \makecell{\textbf{Query from} \\ \textbf{Scene Graph Module}} \\ \midrule \endhead
\hline
\endfoot
\hline
\endlastfoot

         & \makecell{Descriptive} & \makecell{Where is \\the white\\ chair?} & \makecell{chair or \\ armchair}&  \includegraphics[width=3cm]{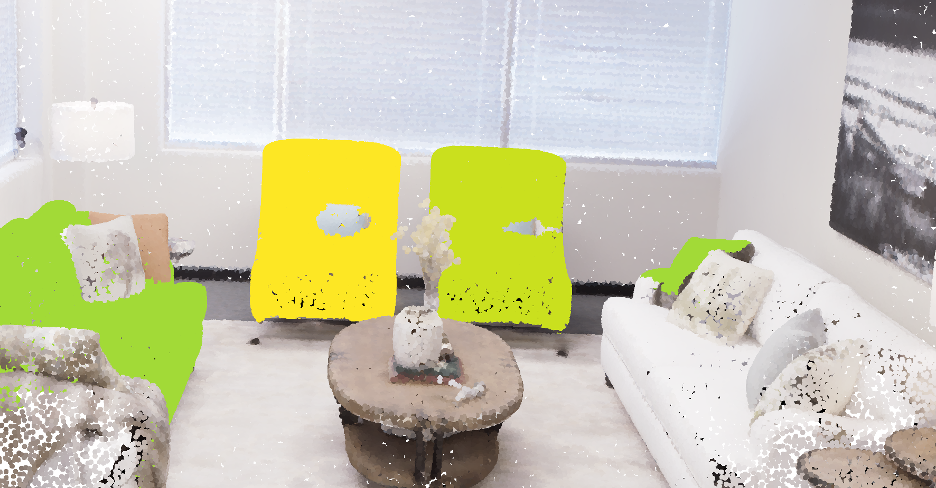}& \includegraphics[width=3cm]{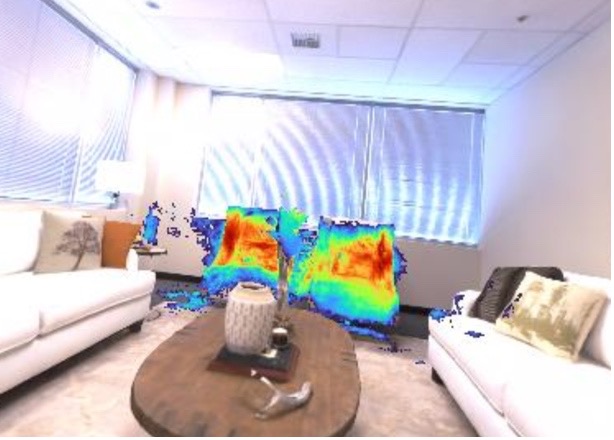}& \includegraphics[width=4cm]{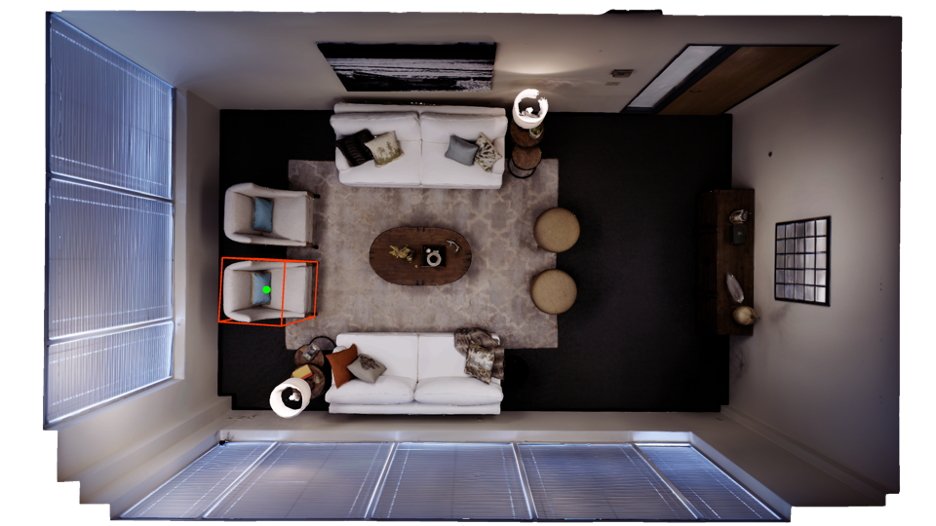}\\ \cline{2-4}
         {\multirow{3}{*}{\makecell{Room 0}}} & \makecell{Descriptive}& \makecell{Anything \\fragile in \\ the room?} & \makecell{ sculpture }   & \includegraphics[width=3cm]{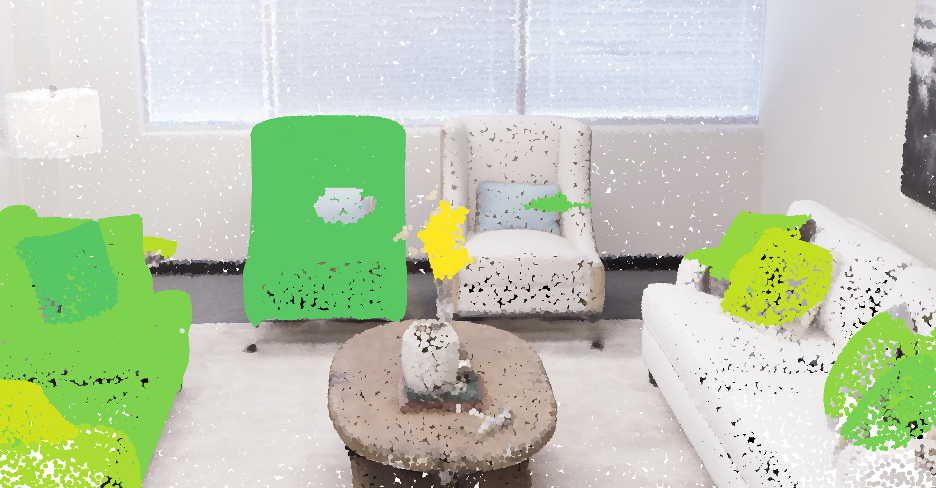} & \includegraphics[width=3cm]{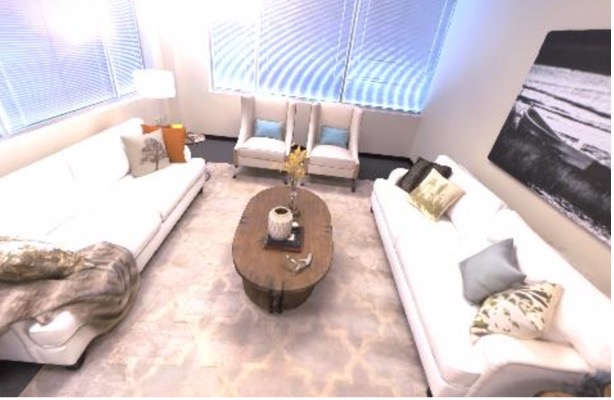} & \includegraphics[width=4cm]{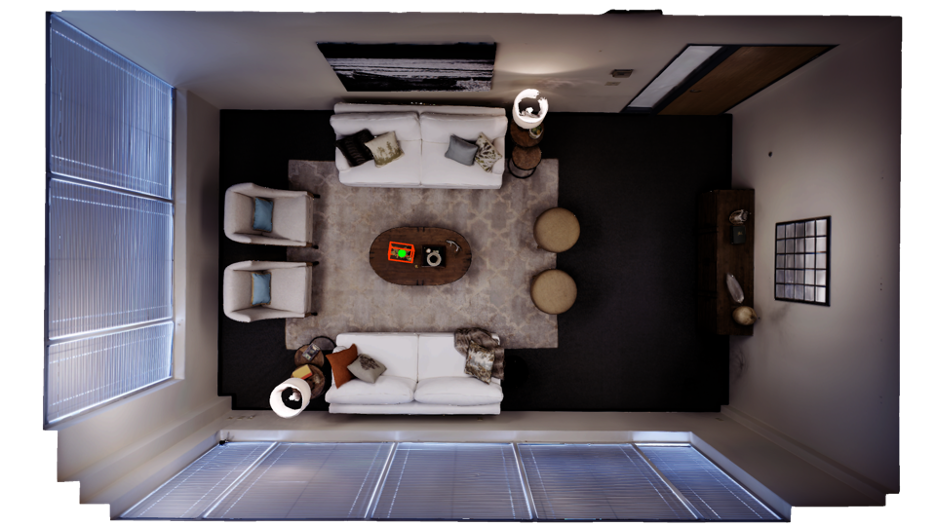}\\ \cline{2-4}
         & \makecell{Descriptive}& \makecell{Where is \\the pillow \\with tree \\pattern on \\it?} & \makecell{pillow} & { \includegraphics[width=3cm]{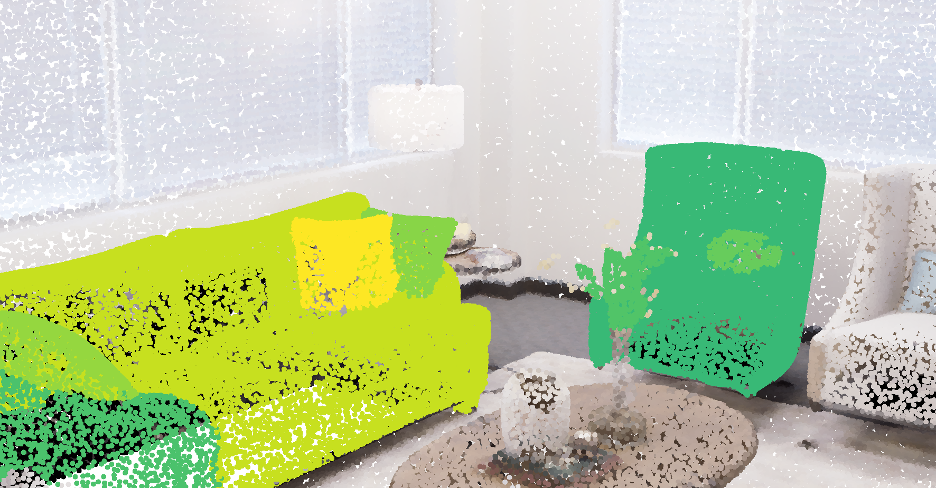}}& { \includegraphics[width=3cm]{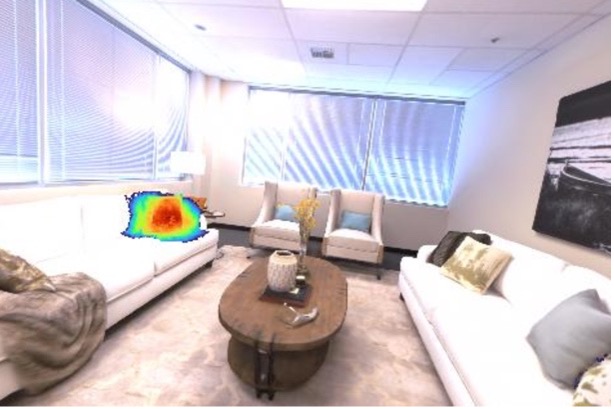}}&{\includegraphics[width=4cm]{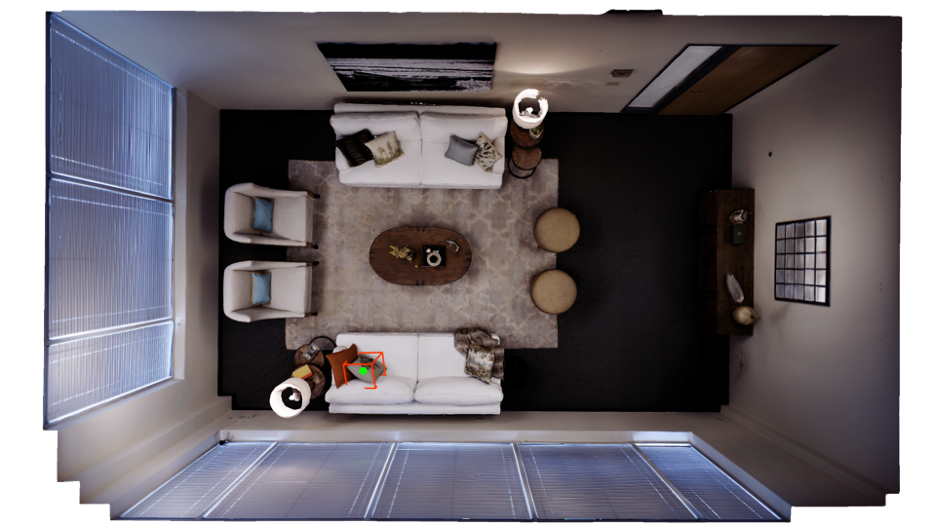}} \\ \midrule

         & \makecell{Affordance}& \makecell{Is there any\\ decoration \\in the room \\such as \\plant?} & \makecell{plant}&{ \includegraphics[width=3cm]{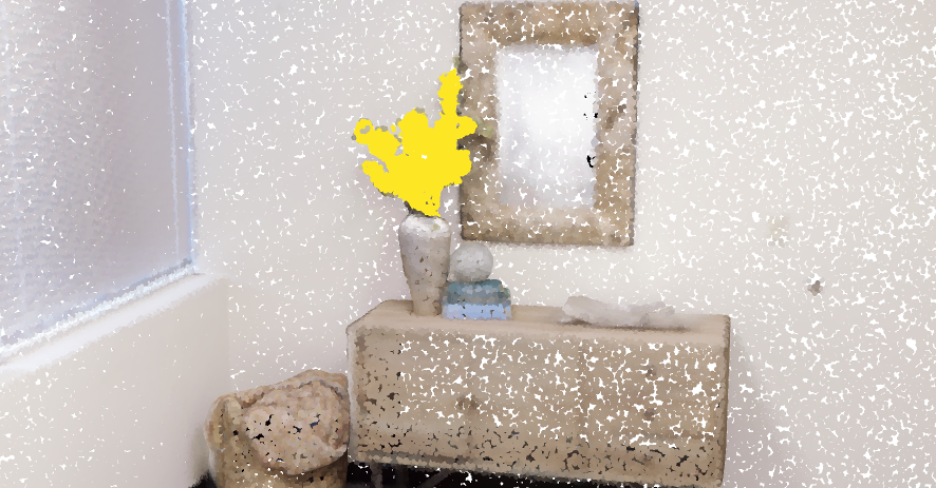}} & { \includegraphics[width=3cm]{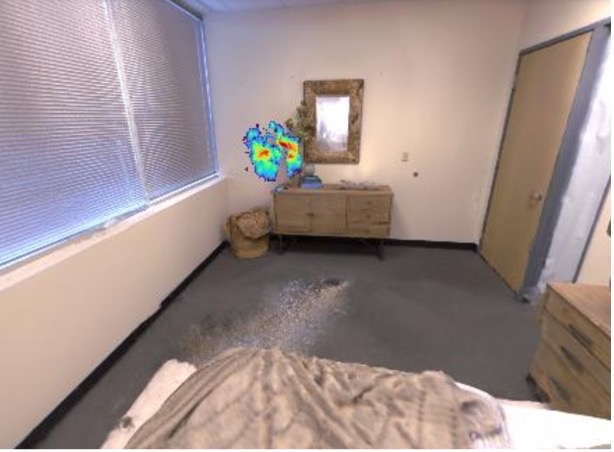}}& {\includegraphics[width=4cm]{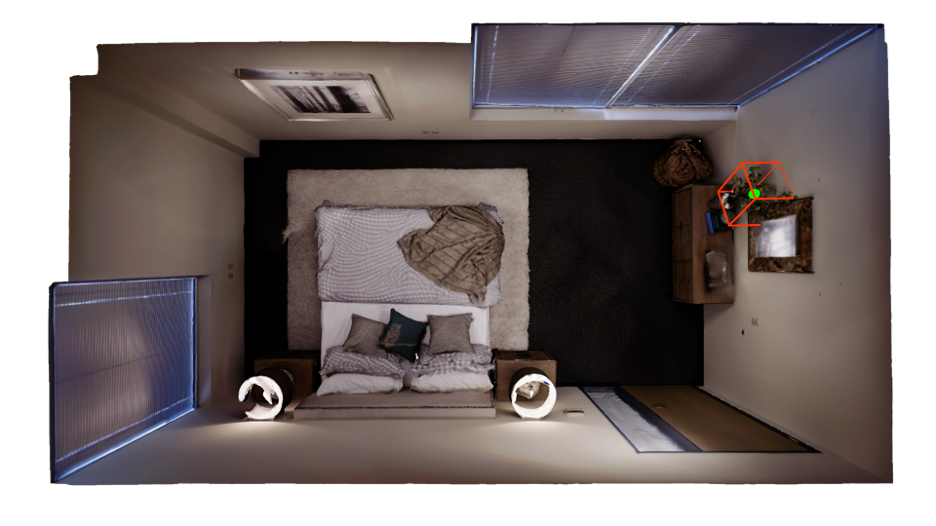}} \\  \cline{2-4}
         {\multirow{3}{*}{Room 1}}
         & \makecell{Descriptive}& \makecell{How many\\ nightstands \\in the\\room?} & \makecell{nightstand}&{ \includegraphics[width=3cm]{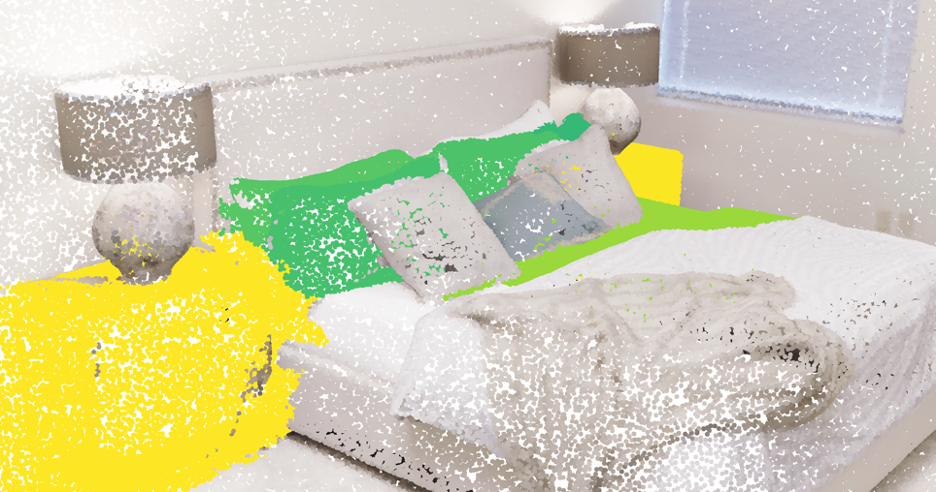}} & { \includegraphics[width=3cm]{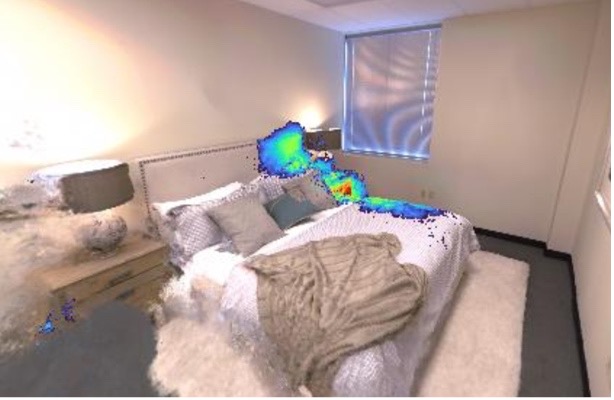}}& {\includegraphics[width=4cm]{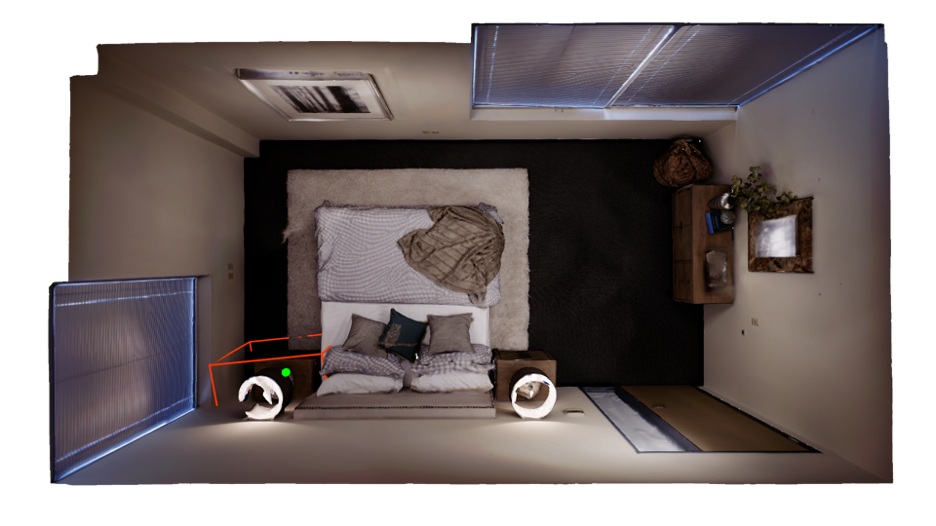}} \\ \cline{2-4}
         & \makecell{Descriptive}& \makecell{What's on\\ the bed?} & \makecell{rug, pillow}&{ \includegraphics[width=3cm]{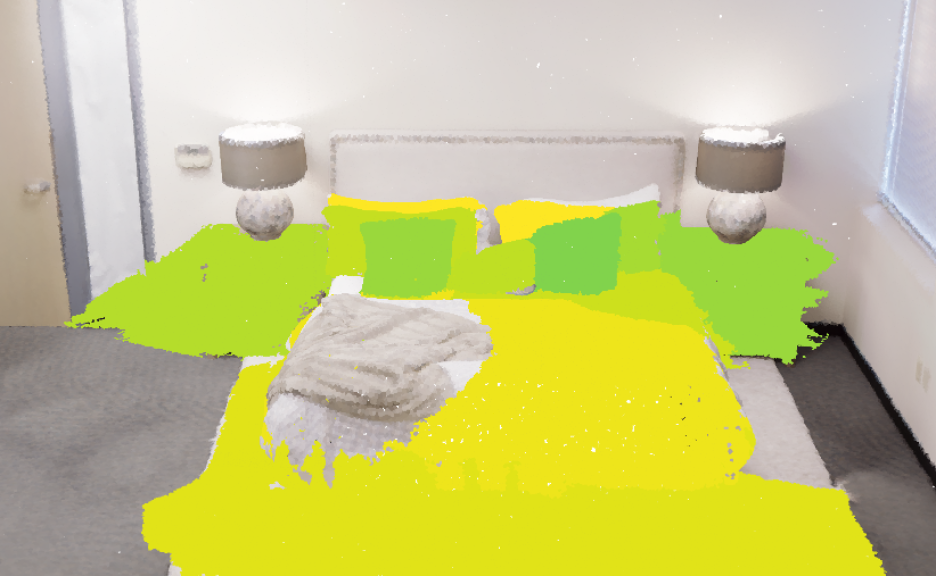} }& { \includegraphics[width=3cm]{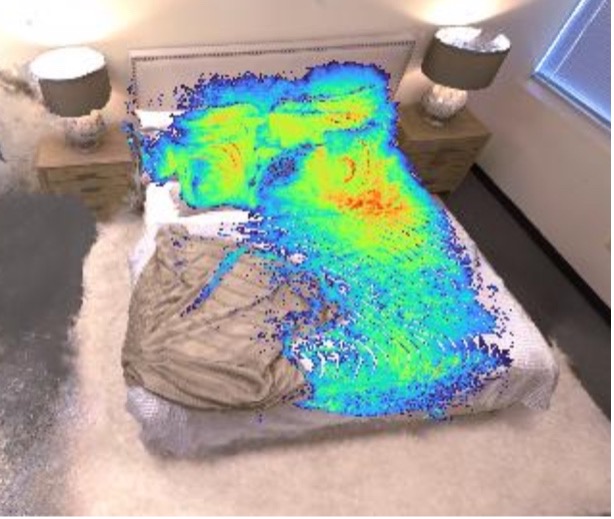}}& \makecell{No relevant objects found.} \\ \midrule

         & \makecell{Descriptive}& \makecell{Where is the\\ vase?} & \makecell{vase}&{ \includegraphics[width=3cm]{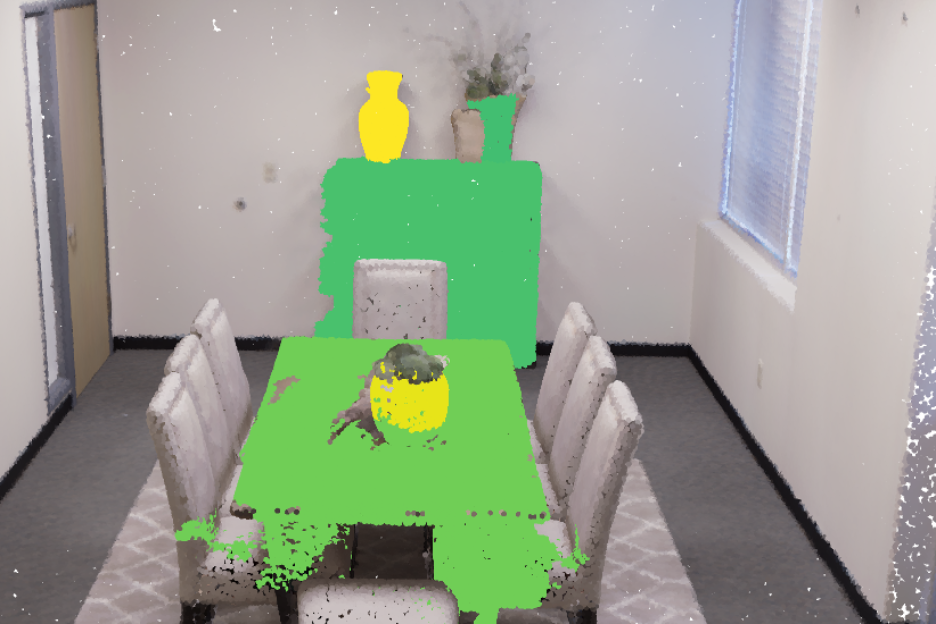}  } & {\includegraphics[width=3cm]{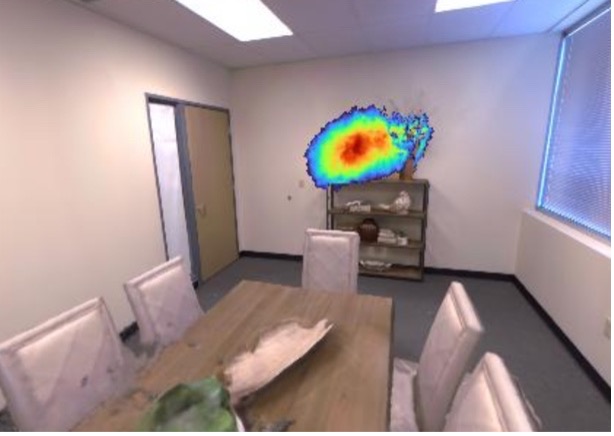}}& {\includegraphics[width=4cm]{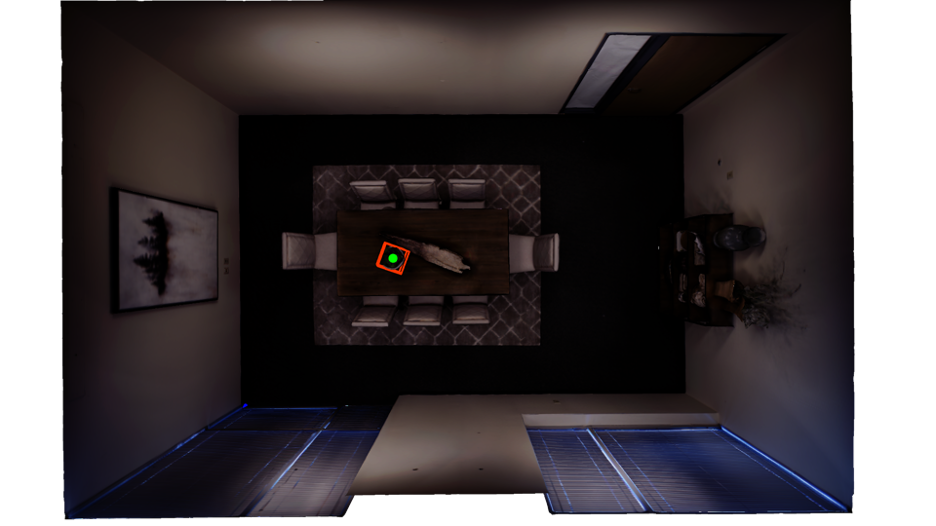}} \\ \cline{2-4}
         {\multirow{3}{*}{Room 2}}& \makecell{Descriptive}& \makecell{How many \\ chairs are \\there in the \\room?} & \makecell{chair}&{  \includegraphics[width=3cm]{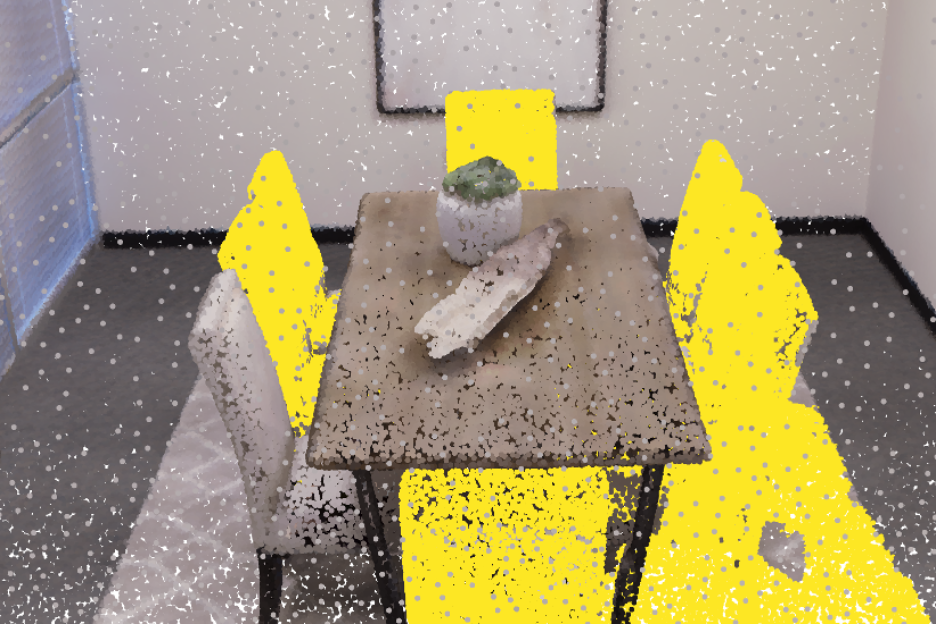}} & { \includegraphics[width=3cm]{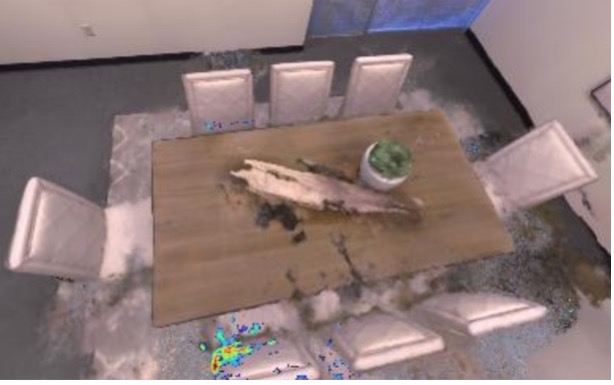}}& {\includegraphics[width=4cm]{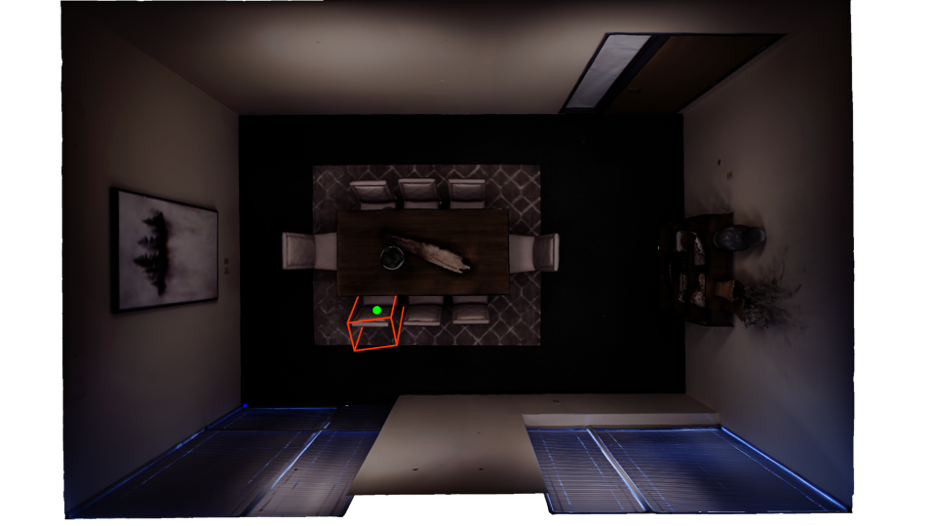}} \\\cline{2-4}
         & \makecell{Affordance}& \makecell{Is there \\anywhere \\to store \\and display \\items in the \\room?} & \makecell{shelf} & {\includegraphics[width=3cm]{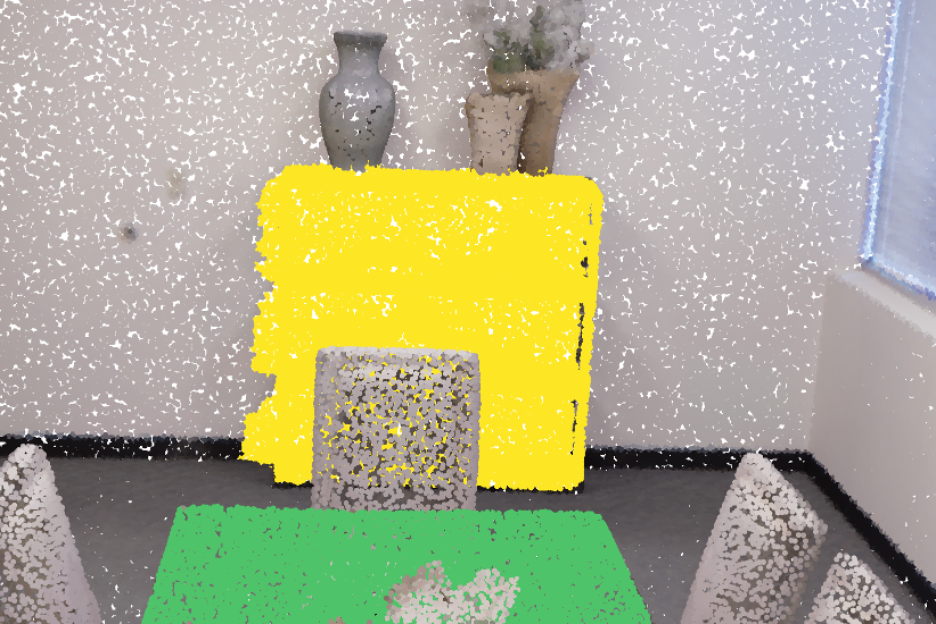}}& {\includegraphics[width=3cm]{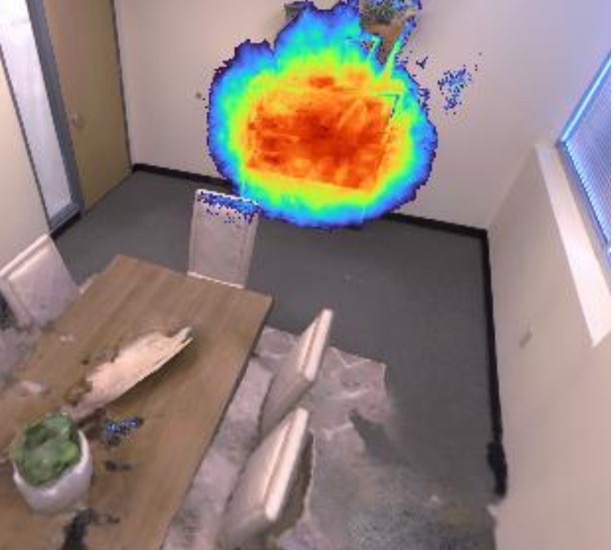}}& \includegraphics[width=4cm]{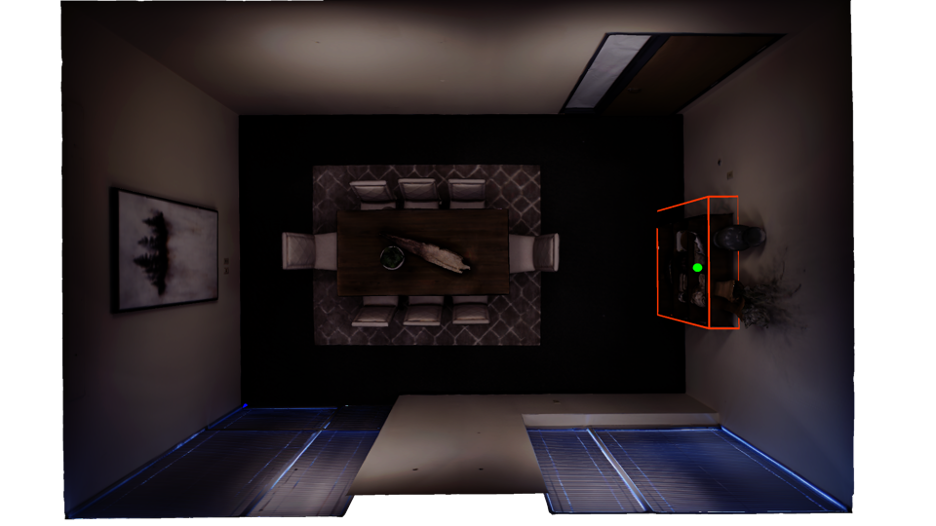} \\ \midrule

& \makecell{Affordance}& \makecell{I need to \\ know the \\ time.} & \makecell{clock}&{ \includegraphics[width=3cm]{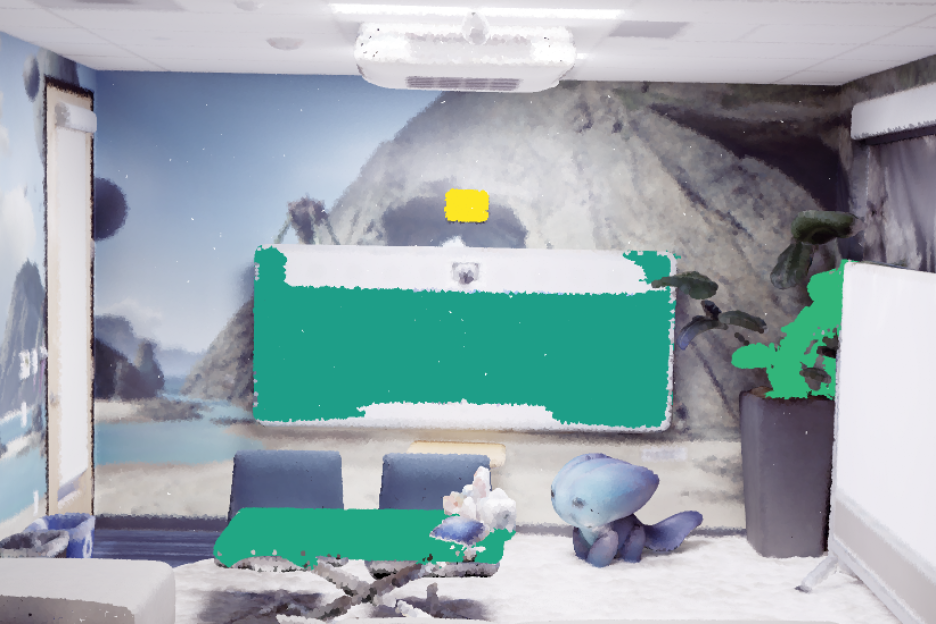}} & { \includegraphics[width=3cm]{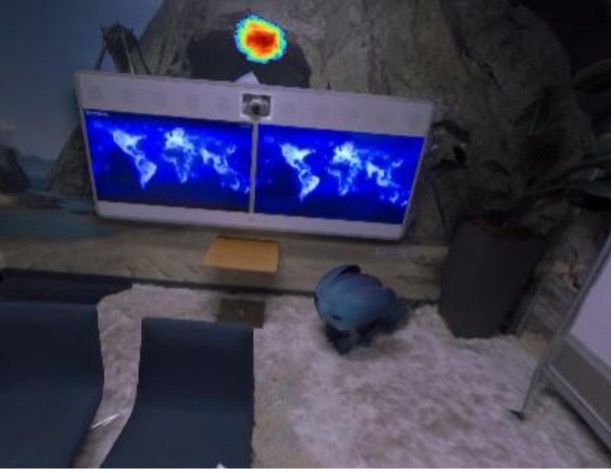}}& \makecell{No relevant objects found.} \\ \cline{2-4}
\multirow{3}{*}{Office 0}& \makecell{Affordance}& \makecell{Does the \\ room have \\ screen for\\ presentation?} & \makecell{television}&{ \includegraphics[width=3cm]{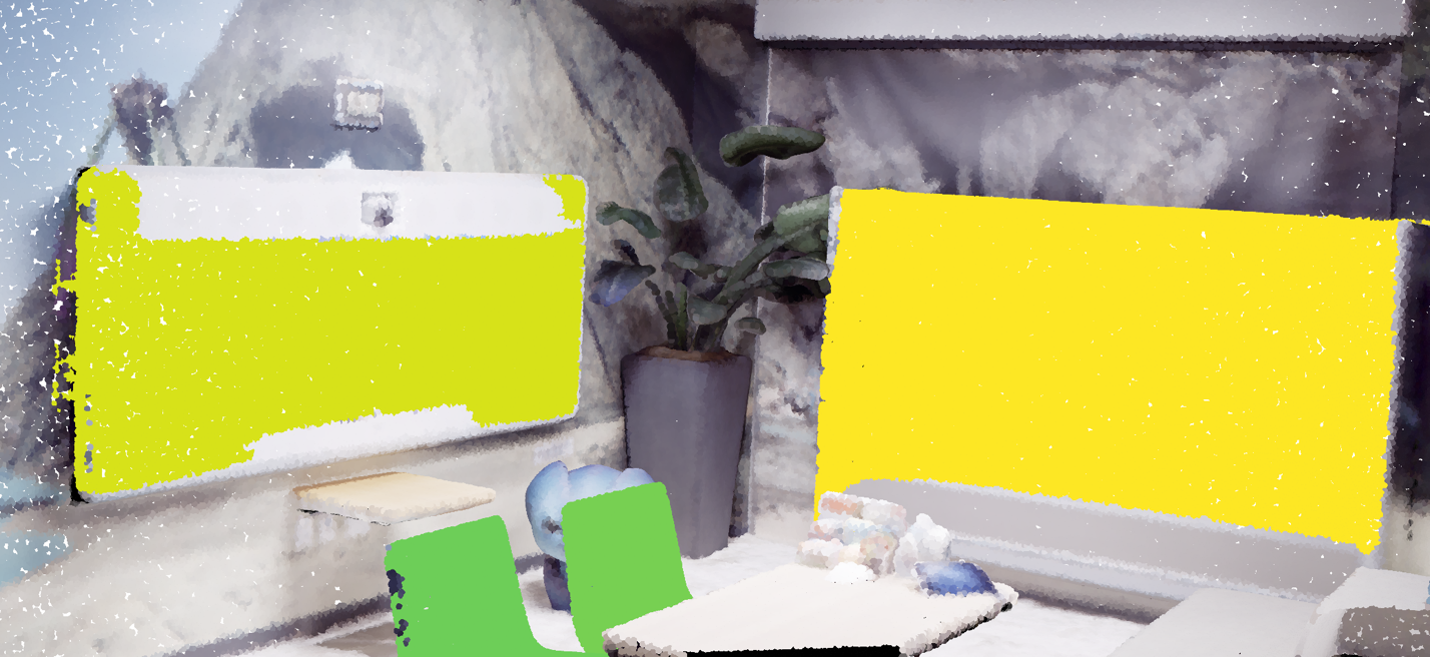}} & { \includegraphics[width=3cm]{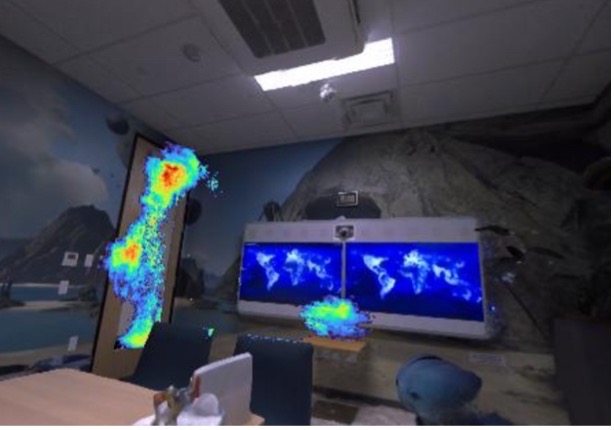}}& {\includegraphics[width=4cm]{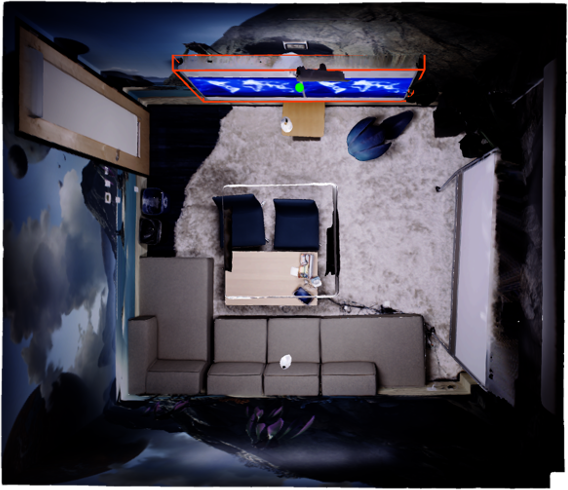}} \\ \cline{2-4}
& \makecell{Descriptive}& \makecell{Any plant\\ decoration \\in the \\ room?} & \makecell{plant }&{ \includegraphics[width=3cm]{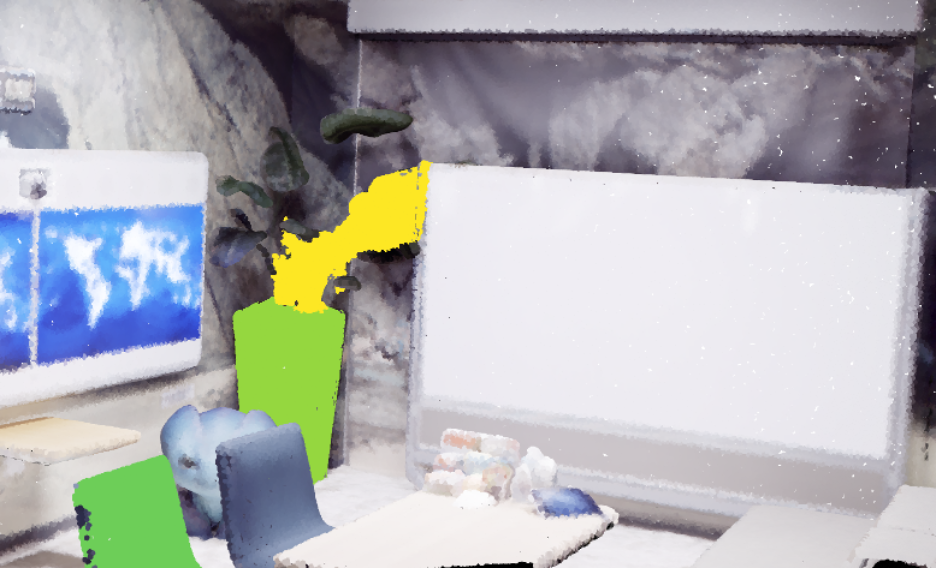}} & { \includegraphics[width=3cm]{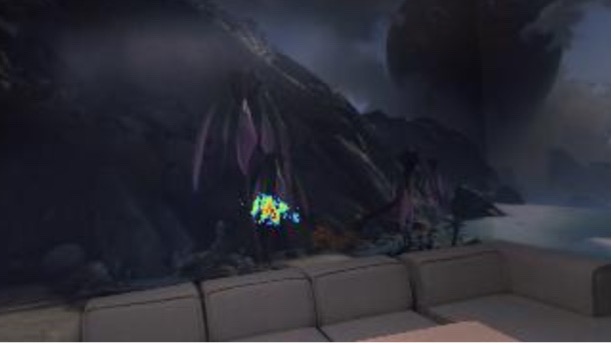}}& {\includegraphics[width=4cm]{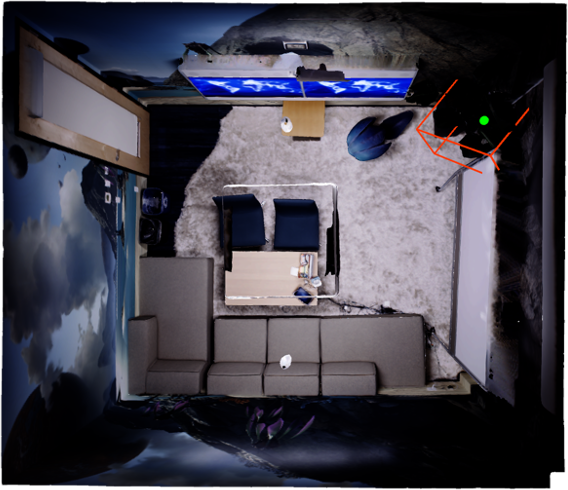}} \\ \midrule

& \makecell{Negation}& \makecell{Anything \\to sit on\\ other than \\a chair?} & \makecell{sofa}&{\includegraphics[width=3cm]{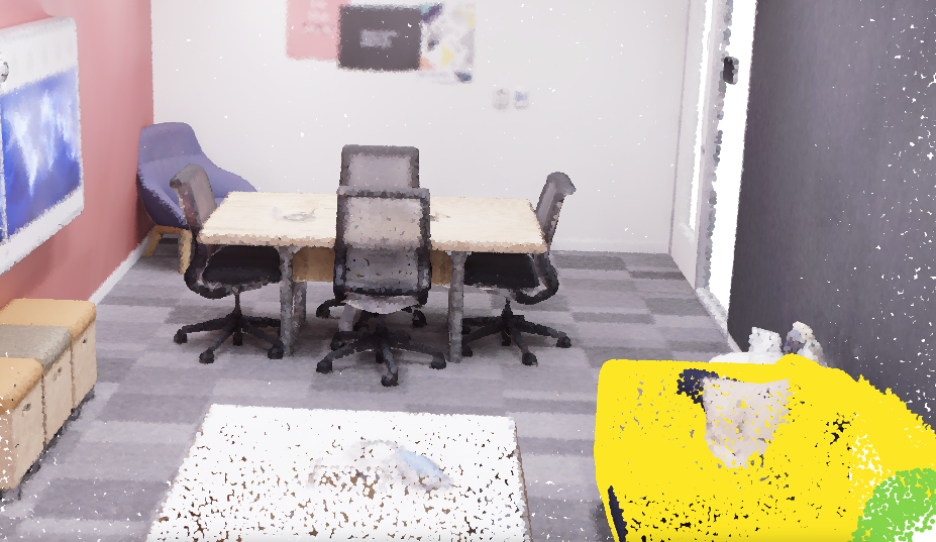}} & { \includegraphics[width=3cm]{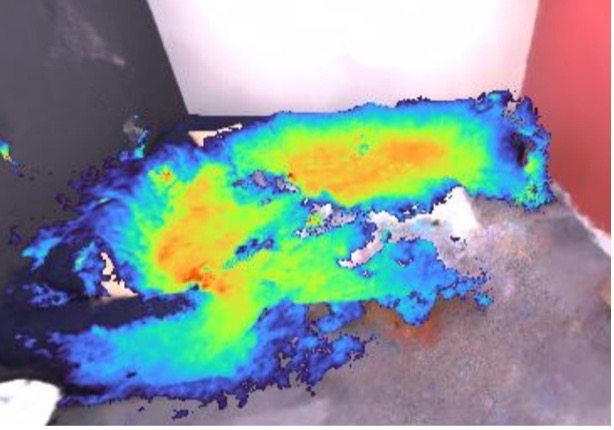}}& \makecell{No relevant objects found.} \\ \cline{2-4}
\multirow{3}{*}{Office 2}& \makecell{Affordance}& \makecell{Somewhere \\ to sit \\ upright for \\ a work call.} & \makecell{chair}&{ \includegraphics[width=3cm]{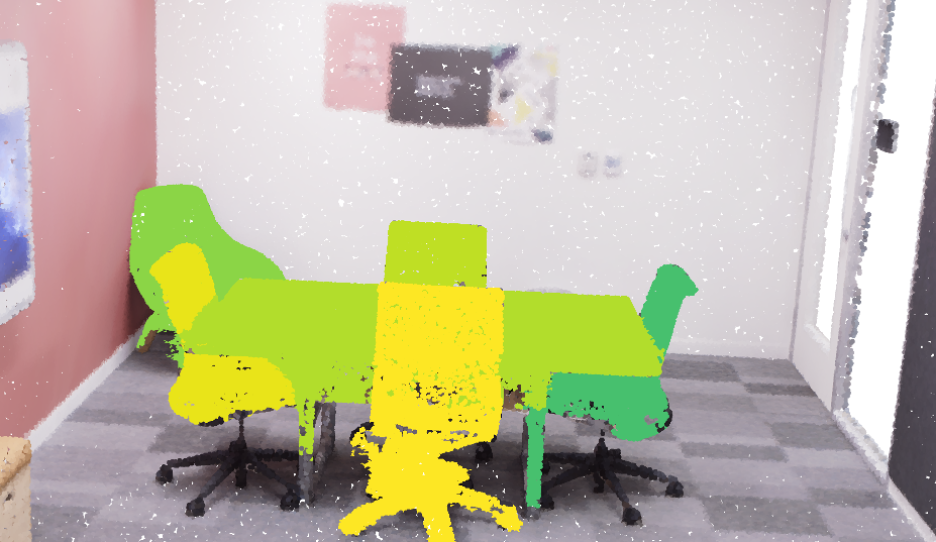}} & { \includegraphics[width=3cm]{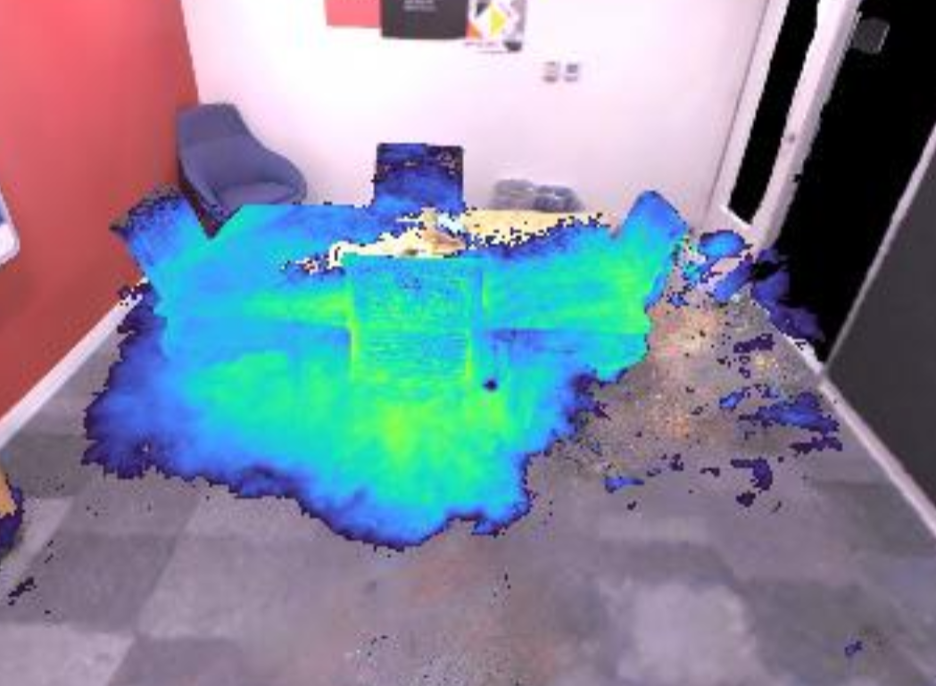}}& {\includegraphics[width=4cm]{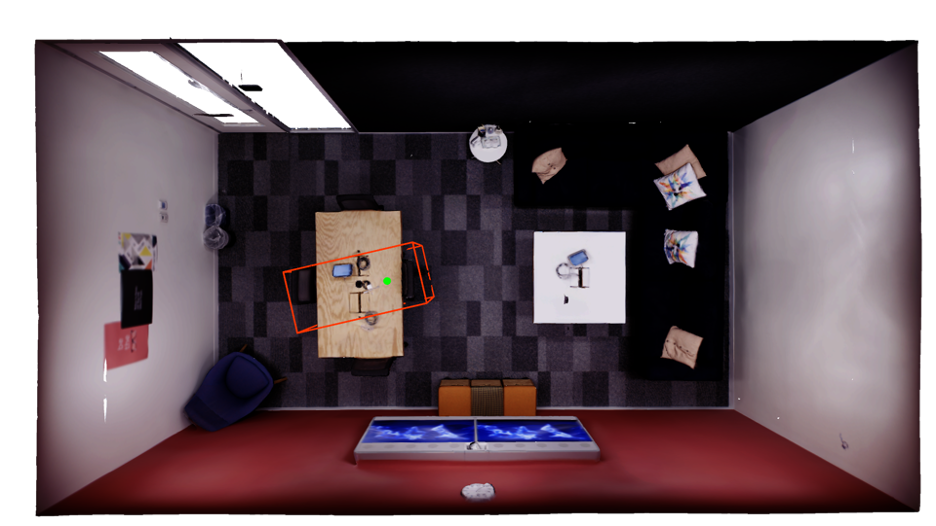}} \\ \cline{2-4}
& \makecell{Affordance}& \makecell{Anything to \\add light to \\the room?} & \makecell{light}&{ \includegraphics[width=3cm]{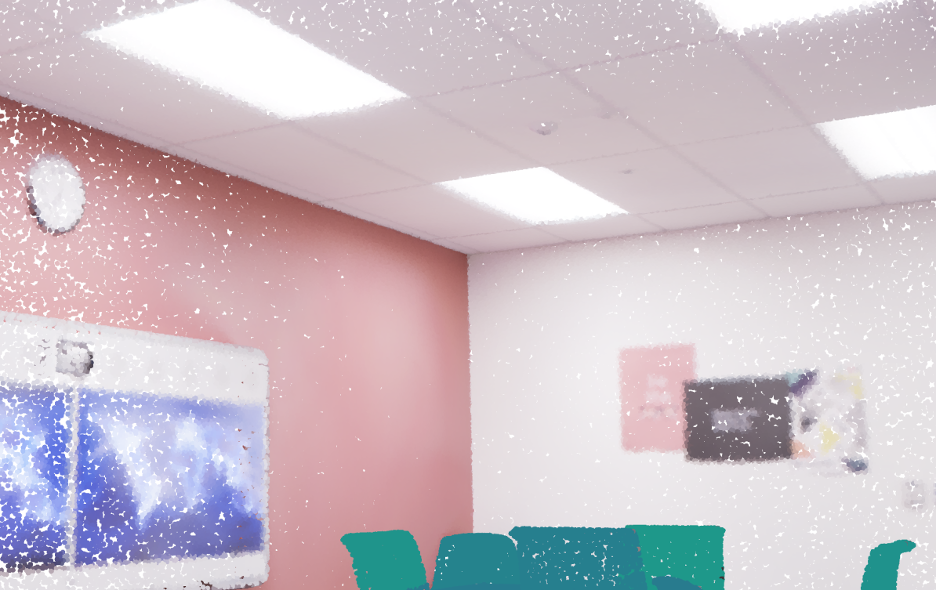}} & { \includegraphics[width=3cm]{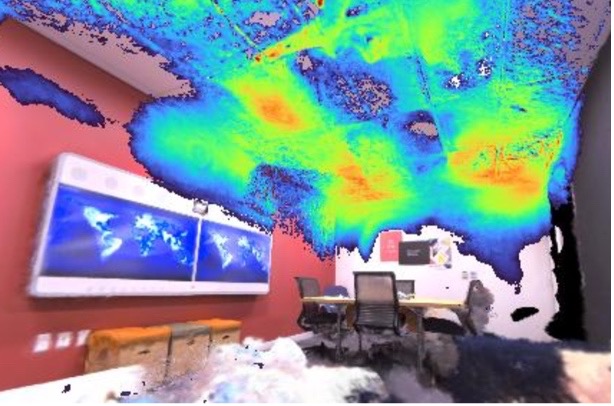}}& \makecell{No relevant objects found.} \\ \midrule

& \makecell{Affordance}& \makecell{Something \\ to sit more\\ than one\\ person.}& \makecell{bench,\\ sofa} &{ \includegraphics[width=3cm]{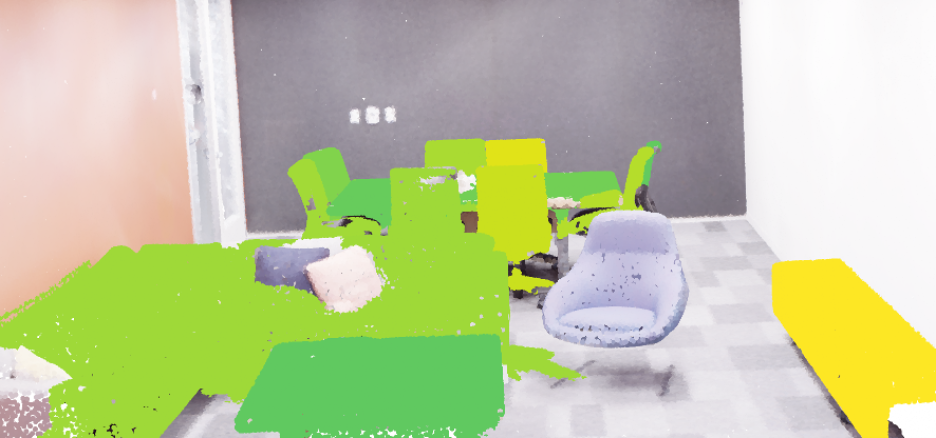}} & { \includegraphics[width=3cm]{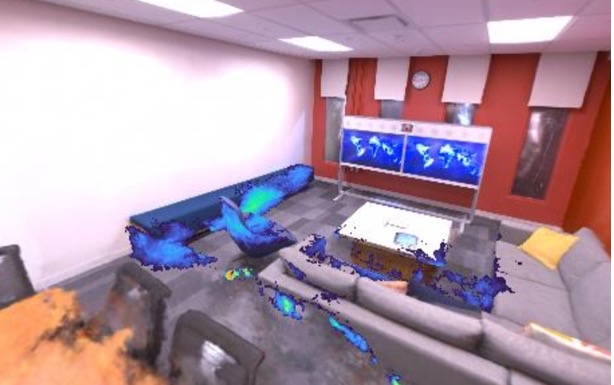}}
& {\includegraphics[width=4cm]{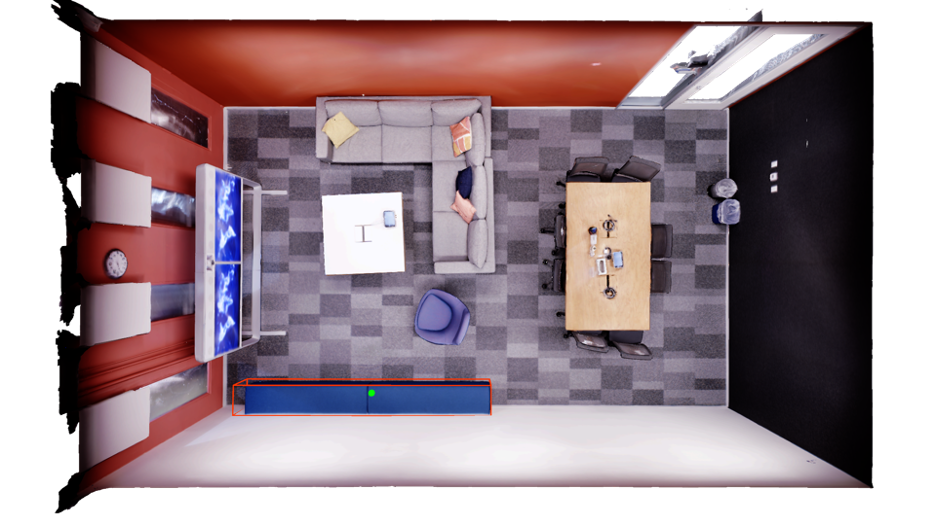}} \\ \cline{2-4}
\multirow{3}{*}{Office 3}& \makecell{Affordance}& \makecell{Somewhere \\to sit and\\ write.} & \makecell{chair}&{ \includegraphics[width=3cm]{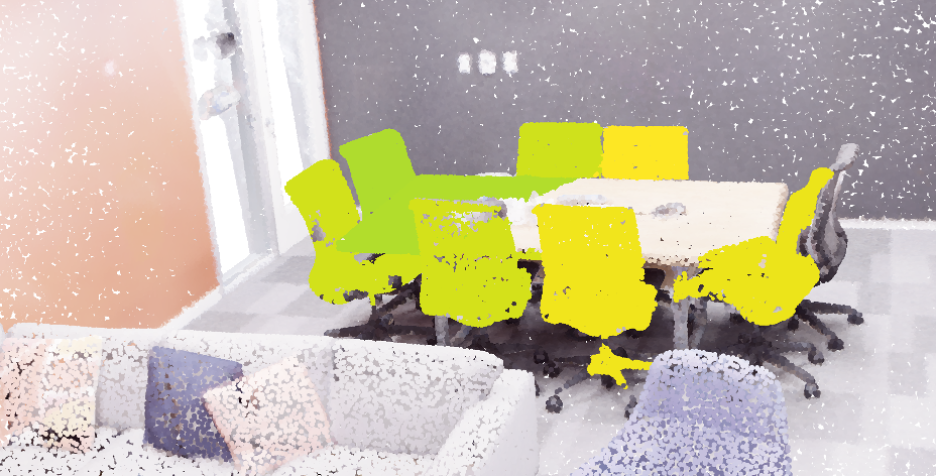}} & { \includegraphics[width=3cm]{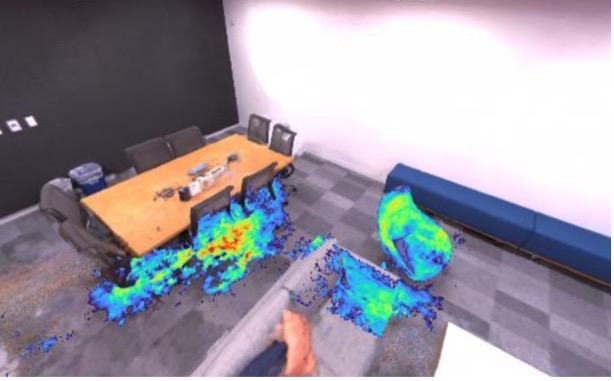}}& {\includegraphics[width=4cm]{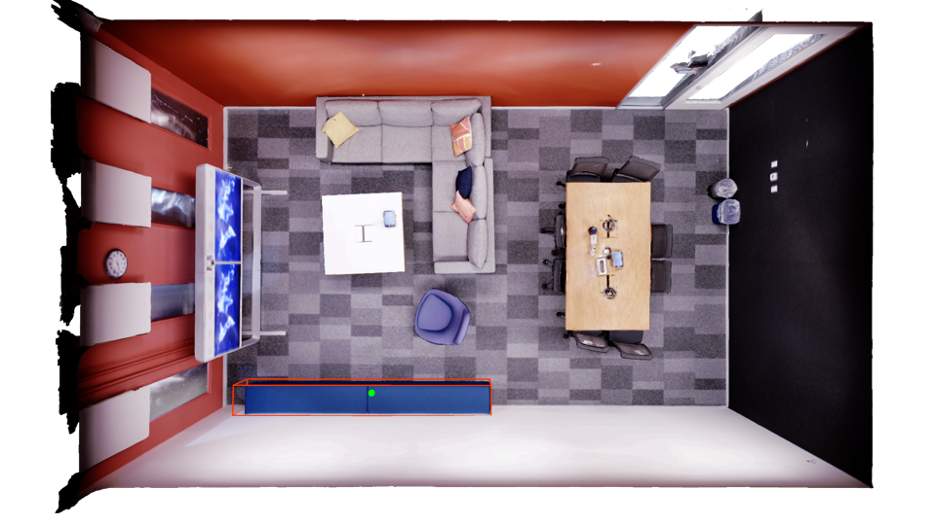}} \\ \cline{2-4}
& \makecell{Affordance}& \makecell{I need to \\ watch TV \\ shows.}& \makecell{television \\ or display \\ unit} &{ \includegraphics[width=3cm]{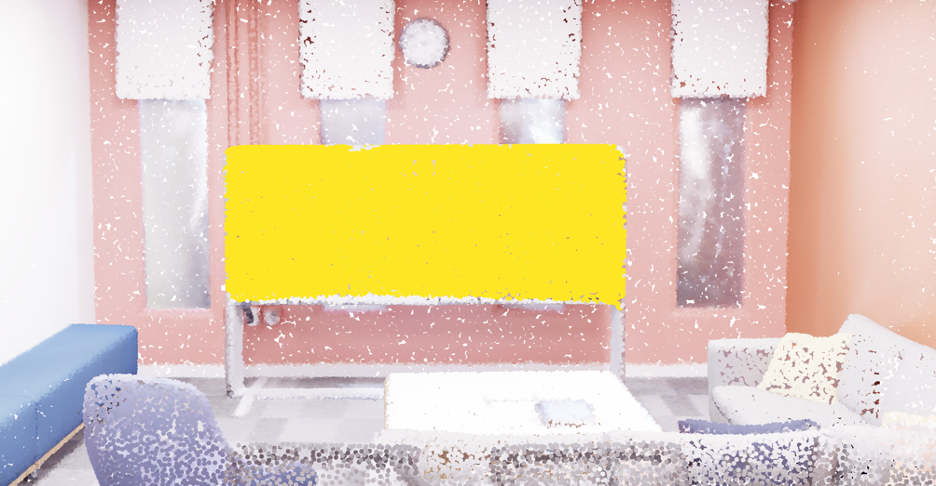}} & { \includegraphics[width=3cm]{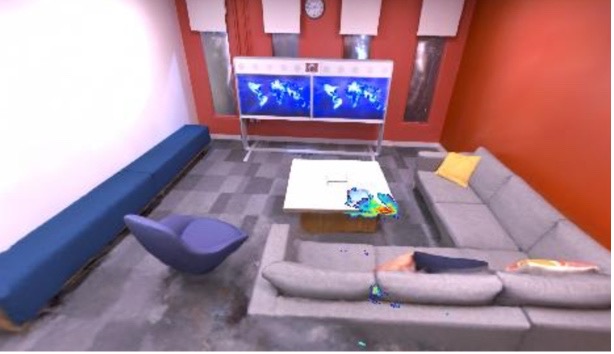}}& {\includegraphics[width=4cm]{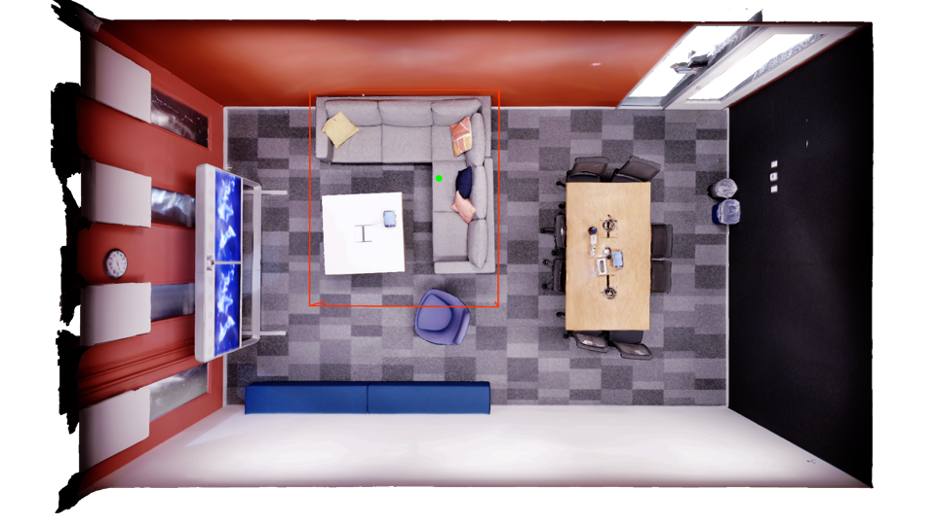}} \\ \bottomrule
    
\end{longtable}

\end{center}

\twocolumn

\section{LLM Generated Queries for Performance Evaluation}
\label{label:appendix-c}

\subsection{Room 0}

\paragraph{Descriptive Queries}
\begin{enumerate}
\item Beige wingback armchair with soft fabric and decorative trim.
\item This is a smooth, white sofa.
\item This is a round, smooth, soft beige ottoman.
\item This is a vibrant yellow book with a red spine.
\item This is a plush white three-seater sofa with pillows.
\item This is a rustic, round, earthy brown side table.
\item This is a gold, brown, silky decorative pillow.
\item Decorative pillow with tree silhouette in cream and brown.
\item Yello-flame like Sculpture on top of a table.
\item This plush throw blanket is rich taupe and muted brown.
\end{enumerate}

\paragraph{Affordance Queries}
\begin{enumerate}
\item Something to enhance comfort on a sofa.
\item Somewhere to sit for relaxation.
\item Something that offers warmth in cold. 
\item Something to rest your feet on.
\item Something to read.
\end{enumerate}

\paragraph{Negation Queries}
\begin{enumerate}
\item Something smooth, unlike a rug. 
\item Something rectangular, unlike a sculpture. 
\item Something soft, unlike a solid table.
\item Something plush, unlike a wooden sideboard. 
\item Something smooth, unlike a textured pillow. 
\end{enumerate}

\subsection{Room 1}
\paragraph{Descriptive Queries}
\begin{enumerate}
\item Subtle gray pillow, soft and textured, for comfort and style.
\item Rustic wooden nightstand with warm tone and textured finish.
\item This is a soft, white bed for sleeping.
\item This is a soft decorative pillow with blue checkered pattern.
\item Luxurious cream rug enhances style and comfort.
\item Muted teal pillow with gold owl and swirling patterns.
\item This is a soft, textured, muted grey decorative pillow.
\item This is a decorative plant in a silver metallic vase.
\item This is a soft, plush, crisp white rectangular pillow.
\item This is a nightstand next to the window.
\end{enumerate}

\paragraph{Affordance Queries}
\begin{enumerate}
\item Something to rest my head on while sleeping.
\item Something that holds a lamp.
\item Something plush to sit on in a chair.
\item Something decorative for my indoor space.
\item Somewhere to rest comfortably at night.
\end{enumerate}
\paragraph{Negation Queries}
\begin{enumerate}
\item Something wooden, unlike a soft pillow.
\item Something rigid, unlike a plush pillow.
\item Something small, unlike a large bed.
\item Something soft, unlike a rough nightstand.
\item Something man-made, unlike a plant.
\end{enumerate}

\subsection{Room 2}
\paragraph{Descriptive Queries}
\begin{enumerate}
\item This is a polished, smooth dining table with earthy tones.
\item Glossy white ceramic vase with small succulent plant.
\item A smooth, muted grey-blue vase with gold speckling.
\item Padded dining chair with diamond stitching, light beige and black.
\item Painting on the wall.
\item Plant in the room.
\item Bird shaped sculpture in the room.
\item This is a black matte industrial-style shelving unit.
\item Grey, soft, woven-patterned dining chair.
\item Chair next to the door.
\end{enumerate}

\paragraph{Affordance Queries}
\begin{enumerate}
\item Something to display decorative items on.
\item Something to sit at a dining table.
\item Something for group meals.
\item Somewhere to hang on the wall.
\item Something to hold flowers.
\end{enumerate}

\paragraph{Negation Queries}
\begin{enumerate}
\item Something wooden, unlike a fragile vase.
\item Somewhere to sit, unlike a table.
\item Something soft to sit on, unlike a hard bench.
\item Something lightweight, unlike a heavy table.
\item Living plant, unlike a chair.
\end{enumerate}

\subsection{Office 0}
\paragraph{Descriptive Queries}
\begin{enumerate}
    \item This is a sleek, dark charcoal grey planter.
    \item This is a smooth black digital desk clock with silver edges.
    \item Bird of paradise, glossy dark green leaves, enhances indoor ambiance.
    \item This is a soft, neutral gray, L-shaped sectional sofa.
    \item Dual-screen video conferencing system with integrated camera.
    \item Sleek, minimalist chair in deep black, versatile seating.
    \item This is a smooth, light wood coffee table for decor.
    \item This is a matte black flat-screen television with a silver border.
    \item This is a deep navy blue, smooth, armless chair.
    \item Large gray displays with a world map, glossy surface.
\end{enumerate}

\paragraph{Affordance Queries}
\begin{enumerate}
    \item Something to add greenery indoors.
    \item Somewhere to seat multiple people comfortably.
    \item Something for ergonomic seating.
    \item Something to conduct video conference with.
    \item Something to tell the time.
\end{enumerate}

\paragraph{Negation Queries}
\begin{enumerate}
    \item Something showing time, unlike rotating dial clock.
    \item Something to sit on, unlike a sofa.
    \item Something wooden, unlike a soft toy.
    \item Something small, unlike a sectional sofa.
    \item Something to watch news on, unlike newspaper.
\end{enumerate}

\subsection{Office 2}
\paragraph{Descriptive Queries}
\begin{enumerate}
\item Deep navy blue plush sofa with light-hued pillows.
\item Deep blue armchair with smooth texture for comfortable seating.
\item Black office chair backrest with grey mesh lines.
\item This is a black mesh office chair with cushioned seat.
\item Sleek white coffee table with wood edge, central piece.
\item Black ergonomic chair with breathable mesh and contoured support.
\item White pillow with multicoloured abstract pattern for decor.
\item This is a plush, deep navy blue sectional sofa.
\item This is a polished conference table with light wood.
\item This is a dual screen.
\end{enumerate}

\paragraph{Affordance Queries}
\begin{enumerate}
\item Something to place decorative items on.
\item Somewhere to sit for ergonomic support.
\item Something to provide comfortable seating for reading.
\item Something to enhance aesthetic appeal to the sofa.
\item Side table by the sofa.
\end{enumerate}
\paragraph{Negation Queries}
\begin{enumerate}
\item Something wooden, unlike a armchair.
\item Something sectional, unlike a single seat.
\item Something flat, unlike a chair.
\item Something plush, unlike rigid screen.
\item Something to sit on, unlike a table.

\end{enumerate}

\subsection{Office 3}
\paragraph{Descriptive Queries}
\begin{enumerate}
\item This is a deep blue, smooth, versatile seating bench.
\item White tabletop coffee table with wood base, holds decor.
\item This is a sleek white dual-screen digital display unit.
\item Rectangular wooden table with warm finish for dining gatherings.
\item L-shaped grey sectional sofa with soft texture and pillows.
\item This is a polished wood conference table for meetings.
\item This is a black ergonomic office chair with mesh.
\item This is a red pillow on the sofa.
\item This is a red wall.
\item A blue armchair.
\end{enumerate}

\paragraph{Affordance Queries}
\begin{enumerate}
\item Something to place books and decor on.
\item Something to display digital maps on.
\item Something to sit on for office work.
\item Something to facilitate meetings.
\item Somewhere for group of coworkers to sit.
\end{enumerate}

\paragraph{Negation Queries}
\begin{enumerate}
\item Something wooden, unlike a display unit.
\item Something ergonomic, unlike a stool.
\item Something soft, unlike a display unit.
\item Something small, unlike a conference table.
\item Something for digital display, unlike a painting.
\end{enumerate}

\end{document}